\RequirePackage{etoolbox}
\newtoggle{icml}
\newtoggle{neurips}

\togglefalse{icml}
\togglefalse{neurips}     
 
\documentclass{article}
 
\newcommand{\icml}[1]{\iftoggle{icml}{#1}{}}
\newcommand{\neurips}[1]{\iftoggle{neurips}{#1}{}}
\newcommand{\arxiv}[1]{\iftoggle{icml}{}{\iftoggle{neurips}{}{#1}}}
\newcommand{\venue}[1]{\iftoggle{icml}{#1}{\iftoggle{neurips}{#1}{}}}

\icml{
  \usepackage{icml2026}
}

\neurips{
  \usepackage[nonatbib]{neurips_2026}   \usepackage{natbib}
  \bibpunct{(}{)}{;}{a}{,}{,}
}

\arxiv{  
	\usepackage{natbib}
	\bibpunct{(}{)}{;}{a}{,}{,}
}

\arxiv{
	\usepackage[letterpaper, left=1in, right=1in, top=1in,
        bottom=1in]{geometry}
        \usepackage[hypertexnames=false]{hyperref}
        \hypersetup{colorlinks,citecolor=blue!70!black,linkcolor =
          blue!70!black}       \usepackage{parskip}
      }

\arxiv{
  \usepackage[dvipsnames]{xcolor}         }
\neurips{
  \usepackage[dvipsnames]{xcolor}
  \usepackage[hypertexnames=false,colorlinks=true,
              citecolor=blue!70!black,linkcolor=blue!70!black,urlcolor=black,breaklinks=true]{hyperref}
}
\icml{
  \PassOptionsToPackage{dvipsnames}{xcolor}
  \PassOptionsToPackage{hypertexnames=false}{hyperref}
  \usepackage[colorlinks=true, linkcolor=blue!70!black, citecolor=blue!70!black,urlcolor=black,breaklinks=true]{hyperref}
}

\usepackage{newtxmath}

\usepackage[utf8]{inputenc} \usepackage[T1]{fontenc}    \usepackage{url}            \usepackage{booktabs}       \usepackage{amsfonts}       \usepackage{nicefrac}       \usepackage{microtype}      \arxiv{\usepackage{amsthm}}
\neurips{\usepackage{amsthm}}

\usepackage{mathtools} 
\usepackage{amsmath}
\usepackage{bm}
\usepackage{bbm}
\usepackage{amssymb} 
\DeclareSymbolFont{upgreek}{U}{eur}{m}{n}
\DeclareMathSymbol{\uppi}{0}{upgreek}{"19}
\usepackage{makecell}
\usepackage{enumitem}
\usepackage{comment}
\usepackage{wrapfig}
\usepackage{colortbl}
\usepackage{setspace}
\usepackage{transparent}
\usepackage{inconsolata}
\usepackage[scaled=.90]{helvet}
\usepackage{xspace}
\usepackage{verbatim}
\usepackage{titletoc}
\usepackage{algorithm}
\usepackage{algorithmic}

\makeatletter
\AtBeginDocument{\@ifundefined{ALG@step}{}{\let\oldALG@step\ALG@step
    \renewcommand{\ALG@step}{\oldALG@step
      \xdef\ALG@currentHref{ALG@line.\thealgorithm.\arabic{ALG@line}}\global\let\@currentHref\ALG@currentHref
      \xdef\@currentlabel{\arabic{ALG@line}}}\patchcmd{\ALG@doentity}{\noindent\hskip\ALG@tlm}{\noindent\hskip\ALG@tlm\Hy@raisedlink{\hypertarget{\ALG@currentHref}{}}}{}{}}}
\makeatother

\usepackage{tablefootnote}
\usepackage{pifont}

\icml{\usepackage[nameinlink,capitalize]{cleveref}}
\neurips{\usepackage[nameinlink,capitalize]{cleveref}}
\arxiv{\usepackage[nameinlink,capitalize]{cleveref}}

\makeatletter
\newcommand{\neutralize}[1]{\expandafter\let\csname c@#1\endcsname\count@}
\makeatother

\arxiv{
\usepackage{thmtools}
\declaretheoremstyle[
  spaceabove=8pt, spacebelow=8pt, headfont=\bfseries,
  bodyfont=\normalfont\itshape,
  headpunct=., notefont=\normalfont\bfseries, 
  postheadspace=0.5em,
]{myspacingstyle}

\declaretheorem[style=myspacingstyle,name=Theorem,parent=section]{theorem}
\declaretheorem[style=myspacingstyle,name=Lemma,parent=section]{lemma}
\declaretheorem[style=myspacingstyle,name=Corollary,parent=section]{corollary}
\declaretheorem[style=myspacingstyle,name=Claim,parent=section]{claim}
\declaretheorem[style=myspacingstyle,name=Assumption, parent=section]{assumption}
\declaretheorem[style=myspacingstyle,name=Condition, parent=section]{condition}

\declaretheorem[style=myspacingstyle,name=Remark,parent=section]{remark}
\declaretheorem[style=myspacingstyle,name=Proposition, parent=section]{proposition}
\declaretheorem[style=myspacingstyle,name=Fact, parent=section]{fact}
\declaretheorem[style=myspacingstyle,name=Definition, parent=section]{definition}
}

\usepackage{aliascnt}
\venue{
\usepackage{amsthm}
\theoremstyle{plain}
\newtheorem{theorem}{Theorem}[section]

\newaliascnt{lemma}{theorem}
\newtheorem{lemma}[lemma]{Lemma}
\aliascntresetthe{lemma}

\newaliascnt{corollary}{theorem}
\newtheorem{corollary}[corollary]{Corollary}
\aliascntresetthe{corollary}

\newaliascnt{proposition}{theorem}
\newtheorem{proposition}[proposition]{Proposition}
\aliascntresetthe{proposition}

\newaliascnt{remark}{theorem}
\theoremstyle{remark}

\aliascntresetthe{remark}

\theoremstyle{definition}

\newaliascnt{assumption}{theorem}
\newtheorem{assumption}[assumption]{Assumption}
\aliascntresetthe{assumption}

\newaliascnt{definition}{theorem}
\newtheorem{definition}[definition]{Definition}
\aliascntresetthe{definition}

\newaliascnt{condition}{theorem}
\newtheorem{condition}[condition]{Condition}
\aliascntresetthe{condition}

\newaliascnt{claim}{theorem}

\aliascntresetthe{claim}

\newaliascnt{fact}{theorem}

\aliascntresetthe{fact}
}

\usepackage{crossreftools}
\AtBeginDocument{\pdfstringdefDisableCommands{\let\Cref\crtCref
    \let\cref\crtcref
  }}

\arxiv{
\makeatletter
\renewenvironment{proof}[1][Proof]{\par\noindent{\bfseries\upshape {#1.}\ }}{\qed\newline}
\makeatother

\theoremstyle{plain}

\usepackage{xpatch}
\xpatchcmd{\proof}{\itshape}{\normalfont\proofnameformat}{}{}
\newcommand{\proofnameformat}{\bfseries}
}

\renewcommand{\eqref}[1]{\texorpdfstring{\hyperref[#1]{(\ref*{#1})}}{(\ref*{#1})}}
\crefformat{equation}{#2Eq.\,(#1)#3}
\Crefformat{equation}{#2Eq.\,(#1)#3}

\Crefformat{figure}{#2Figure~#1#3}
\Crefformat{assumption}{#2Assumption~#1#3}

\Crefname{assumption}{Assumption}{Assumptions}
\crefname{assumption}{Assumption}{Assumptions}
\Crefname{condition}{Condition}{Conditions}
\crefname{condition}{Condition}{Conditions}
\Crefname{claim}{Claim}{Claims}
\crefname{claim}{Claim}{Claims}
\Crefname{fact}{Fact}{Facts}
\crefname{fact}{Fact}{Facts}
\Crefname{definition}{Definition}{Definitions}
\crefname{definition}{Definition}{Definitions}
\Crefname{line}{Line}{Lines}
\crefname{line}{Line}{Lines}

\def\ddefloop#1{\ifx\ddefloop#1\else\ddef{#1}\expandafter\ddefloop\fi}
\def\ddef#1{\expandafter\def\csname bb#1\endcsname{\ensuremath{\mathbb{#1}}}}
\ddefloop ABCDEFGHIJKLMNOPQRSTUVWXYZ\ddefloop
\def\ddefloop#1{\ifx\ddefloop#1\else\ddef{#1}\expandafter\ddefloop\fi}
\def\ddef#1{\expandafter\def\csname b#1\endcsname{\ensuremath{\mathbf{#1}}}}
\ddefloop ABCDEFGHIJKLMNOPQRSTUVWXYZ\ddefloop
\def\ddef#1{\expandafter\def\csname sf#1\endcsname{\ensuremath{\mathsf{#1}}}}
\ddefloop ABCDEFGHIJKLMNOPQRSTUVWXYZ\ddefloop
\def\ddef#1{\expandafter\def\csname c#1\endcsname{\ensuremath{\mathcal{#1}}}}
\ddefloop ABCDEFGHIJKLMNOPQRSTUVWXYZ\ddefloop
\def\ddef#1{\expandafter\def\csname h#1\endcsname{\ensuremath{\widehat{#1}}}}
\ddefloop ABCDEFGHIJKLMNOPQRSTUVWXYZ\ddefloop
\def\ddef#1{\expandafter\def\csname hc#1\endcsname{\ensuremath{\widehat{\mathcal{#1}}}}}
\ddefloop ABCDEFGHIJKLMNOPQRSTUVWXYZ\ddefloop
\def\ddef#1{\expandafter\def\csname t#1\endcsname{\ensuremath{\widetilde{#1}}}}
\ddefloop ABCDEFGHIJKLMNOPQRSTUVWXYZ\ddefloop
\def\ddef#1{\expandafter\def\csname tc#1\endcsname{\ensuremath{\widetilde{\mathcal{#1}}}}}
\ddefloop ABCDEFGHIJKLMNOPQRSTUVWXYZ\ddefloop
\def\ddefloop#1{\ifx\ddefloop#1\else\ddef{#1}\expandafter\ddefloop\fi}
\def\ddef#1{\expandafter\def\csname scr#1\endcsname{\ensuremath{\mathscr{#1}}}}
\ddefloop ABCDEFGHIJKLMNOPQRSTUVWXYZ\ddefloop

\def\ddef#1{\expandafter\def\csname b#1\endcsname{\ensuremath{\mb{#1}}}}
\ddefloop abcdeghijklmnopqrstuvwxyz\ddefloop

\newcommand{\mb}[1]{\boldsymbol{#1}}

\makeatletter
\let\save@mathaccent\mathaccent
\def\if@single#1#2#3{\setbox0\hbox{${\mathaccent"0362{#1}}^H$}\setbox2\hbox{${\mathaccent"0362{\kern0pt#1}}^H$}\ifdim\ht0=\ht2 #3\else #2\fi
}
\def\rel@kern#1{\kern#1\dimexpr\macc@kerna}
\def\widebar#1{\@ifnextchar^{{\wide@bar{#1}{0}}}{\wide@bar{#1}{1}}}
\def\underbar#1{\@ifnextchar_{{\under@bar{#1}{0}}}{\under@bar{#1}{1}}}
\def\wide@bar#1#2{\if@single{#1}{\wide@bar@{#1}{#2}{1}}{\wide@bar@{#1}{#2}{2}}}
\def\under@bar#1#2{\if@single{#1}{\under@bar@{#1}{#2}{1}}{\under@bar@{#1}{#2}{2}}}
\def\wide@bar@#1#2#3{\begingroup
	\def\mathaccent##1##2{\let\mathaccent\save@mathaccent
		\if#32 \let\macc@nucleus\first@char \fi
		\setbox\z@\hbox{$\macc@style{\macc@nucleus}_{}$}\setbox\tw@\hbox{$\macc@style{\macc@nucleus}{}_{}$}\dimen@\wd\tw@
		\advance\dimen@-\wd\z@
		\divide\dimen@ 3
		\@tempdima\wd\tw@
		\advance\@tempdima-\scriptspace
		\divide\@tempdima 10
		\advance\dimen@-\@tempdima
		\ifdim\dimen@>\z@ \dimen@0pt\fi
		\rel@kern{0.6}\kern-\dimen@
		\if#31
		\overline{\rel@kern{-0.6}\kern\dimen@\macc@nucleus\rel@kern{0.4}\kern\dimen@}\advance\dimen@0.4\dimexpr\macc@kerna
		\let\final@kern#2\ifdim\dimen@<\z@ \let\final@kern1\fi
		\if\final@kern1 \kern-\dimen@\fi
		\else
		\overline{\rel@kern{-0.6}\kern\dimen@#1}\fi
	}\macc@depth\@ne
	\let\math@bgroup\@empty \let\math@egroup\macc@set@skewchar
	\mathsurround\z@ \frozen@everymath{\mathgroup\macc@group\relax}\macc@set@skewchar\relax
	\let\mathaccentV\macc@nested@a
	\if#31
	\macc@nested@a\relax111{#1}\else
	\def\gobble@till@marker##1\endmarker{}\futurelet\first@char\gobble@till@marker#1\endmarker
	\ifcat\noexpand\first@char A\else
	\def\first@char{}\fi
	\macc@nested@a\relax111{\first@char}\fi
	\endgroup
}
\def\under@bar@#1#2#3{\begingroup
	\def\mathaccent##1##2{\let\mathaccent\save@mathaccent
		\if#32 \let\macc@nucleus\first@char \fi
		\setbox\z@\hbox{$\macc@style{\macc@nucleus}_{}$}\setbox\tw@\hbox{$\macc@style{\macc@nucleus}{}_{}$}\dimen@\wd\tw@
		\advance\dimen@-\wd\z@
		\divide\dimen@ 3
		\@tempdima\wd\tw@
		\advance\@tempdima-\scriptspace
		\divide\@tempdima 10
		\advance\dimen@-\@tempdima
		\ifdim\dimen@>\z@ \dimen@0pt\fi
		\rel@kern{0.6}\kern-\dimen@
		\if#31
		\underline{\rel@kern{-0.6}\kern\dimen@\macc@nucleus\rel@kern{0.4}\kern\dimen@}\advance\dimen@0.4\dimexpr\macc@kerna
		\let\final@kern#2\ifdim\dimen@<\z@ \let\final@kern1\fi
		\if\final@kern1 \kern-\dimen@\fi
		\else
		\underline{\rel@kern{-0.6}\kern\dimen@#1}\fi
	}\macc@depth\@ne
	\let\math@bgroup\@empty \let\math@egroup\macc@set@skewchar
	\mathsurround\z@ \frozen@everymath{\mathgroup\macc@group\relax}\macc@set@skewchar\relax
	\let\mathaccentV\macc@nested@a
	\if#31
	\macc@nested@a\relax111{#1}\else
	\def\gobble@till@marker##1\endmarker{}\futurelet\first@char\gobble@till@marker#1\endmarker
	\ifcat\noexpand\first@char A\else
	\def\first@char{}\fi
	\macc@nested@a\relax111{\first@char}\fi
	\endgroup
}
\makeatother

\usepackage{graphicx}
\usepackage{subfigure} 
\usepackage{multirow}

\usepackage{array}
\usepackage{longtable} 

\usepackage{todonotes} 
\usepackage{needspace}
\usepackage{changepage} \usepackage{etoc}

\makeatletter
\newcommand{\mline}[1]{\addtocounter{ALC@line}{-1}\begingroup
\def\theALC@line{Algorithm~\thealgorithm, Line~\arabic{ALC@line}}\refstepcounter{ALC@line}\label{#1}\endgroup
}

\crefformat{ALC@line}{#2#1#3}
\Crefformat{ALC@line}{#2#1#3}
\crefformat{line}{#2#1#3}
\Crefformat{line}{#2#1#3}
\makeatother

\graphicspath{{figures/}}

\renewcommand{\epsilon}{\varepsilon}

\renewcommand{\bar}{\overline}

\let\originalleft\left
\let\originalright\right
\renewcommand{\left}{\mathopen{}\mathclose\bgroup\originalleft}
\renewcommand{\right}{\aftergroup\egroup\originalright}

\newcommand{\outersteps}{\textsc{outer steps}\xspace}
\newcommand{\innersteps}{\textsc{inner steps}\xspace}
\newcommand{\rollback}{\textsc{rollback}\xspace}

\newcommand{\alive}{\textsc{alive}\xspace}
\newcommand{\dead}{\textsc{dead}\xspace}
\newcommand{\anc}{\textsc{anc}\xspace}
\newcommand{\winner}{\textsc{winner}\xspace}

\newcommand{\parent}{\textsc{parent}\xspace}
\newcommand{\addchild}{\textsc{AddChild}\xspace}
\newcommand{\step}{\textsc{Step}\xspace}
\newcommand{\evolvegroup}{\textsc{EvolveGroup}\xspace}
\newcommand{\getancestor}{\textsc{GetAncestor}\xspace}
\newcommand{\reset}{\textsc{Reset}\xspace}

\newcommand{\gowu}{\textsc{GowU}\xspace}

\renewcommand{\paragraph}[1]{\par\noindent\textbf{#1.}}

\newcommand{\algcommentlight}[1]{\textcolor{blue!70!black}{\transparent{0.5}\footnotesize{\texttt{\textbf{//\hspace{2pt}#1}}}}}

\newcommand{\algcommentbiglight}[1]{\textcolor{blue!70!black}{\transparent{0.5}\footnotesize{\texttt{\textbf{/* #1~*/}}}}}

\crefname{remark}{Remark}{Remarks}
\Crefname{remark}{Remark}{Remarks}

\usepackage{color-edits}
\definecolor{ForestGreen}{rgb}{0.13, 0.55, 0.13}
\addauthor{zm}{ForestGreen}
\addauthor{jc}{red}

\newcommand{\gwtw}{\textsc{GWTW}\xspace}
\newcommand{\rnd}{\textsc{RND}\xspace}
\newcommand{\goexplore}{\textsc{Go-Explore}\xspace}

\newcommand{\lexargmax}{\operatorname{lex-argmax}}
\newcommand{\lex}{\mathrm{lex}}
\newcommand{\pitfall}{\emph{Pitfall!}\xspace}
\newcommand{\montezuma}{\textit{Montezuma's Revenge}\xspace}
\newcommand{\venture}{\textit{Venture}\xspace}

\icml{\icmltitlerunning{Uncertainty Guided Tree Search for Hard Exploration}}

\begin{document}
\etocdepthtag.toc{main}
\icml{
\twocolumn[
  \icmltitle{Decoupling Exploration and Policy Optimization: \texorpdfstring{\\}{} Uncertainty Guided Tree Search for Hard Exploration}

  \icmlsetsymbol{equal}{*}

  \begin{icmlauthorlist}
    \icmlauthor{Zakaria Mhammedi}{goog}
    \icmlauthor{James Cohan}{goog}
  \end{icmlauthorlist}

  \icmlaffiliation{goog}{Google Research, NYC}

  \icmlcorrespondingauthor{Zakaria Mhammedi}{mhammedi@google.com}

  \icmlkeywords{Reinforcement Learning, Exploration, Epistemic Uncertainty, Self-Supervised Learning}

  \vskip 0.3in
]
 
\printAffiliationsAndNotice{\icmlEqualContribution}
}

\neurips{
\title{Decoupling Exploration and Policy Optimization: \texorpdfstring{\\}{} Uncertainty Guided Tree Search for Hard Exploration}
\author{Anonymous Author(s)
}
\maketitle
}
 
\arxiv{
\title{Decoupling Exploration and Policy Optimization: \\ Uncertainty Guided Tree Search for Hard Exploration}
\author{Zakaria Mhammedi \\
  Google Research, NYC \\
  \texttt{mhammedi@google.com}
  \and
  James Cohan \\
  Google Research, NYC \\
  \texttt{jamesfcohan@google.com}
}
\maketitle
}

\begin{abstract}
    The process of discovery requires active exploration---the act of collecting new and informative data. However, efficient autonomous exploration remains a major unsolved problem. The dominant paradigm addresses this challenge by using Reinforcement Learning (RL) to train agents with intrinsic motivation, maximizing a composite objective of extrinsic and intrinsic rewards. We suggest that this approach incurs unnecessary overhead: while policy optimization is necessary for precise task execution, employing such machinery solely to expand state coverage may be inefficient. In this paper, we propose a new approach that explicitly decouples exploration from policy optimization and bypasses RL entirely during the exploration phase. Our method uses a tree-search strategy inspired by the Go-With-The-Winner algorithm, paired with a measure of uncertainty to systematically drive exploration. By removing the overhead of policy optimization, our approach explores an order of magnitude more efficiently than standard intrinsic motivation baselines on hard exploration benchmarks. Further, we demonstrate that the trajectories discovered during exploration can be distilled into deployable policies using existing supervised backward learning algorithms, achieving state-of-the-art performance by a wide margin on \montezuma, \pitfall, and \venture without relying on domain-specific knowledge. Finally, we demonstrate the generality of our framework in high-dimensional continuous action spaces by solving the MuJoCo Adroit dexterous manipulation and AntMaze tasks in a \emph{sparse-reward} setting, directly from image observations and without expert demonstrations or offline datasets. To the best of our knowledge, this has not been achieved before\arxiv{ for the Adroit tasks}.
\end{abstract}
 
\section{Introduction}
\label{sec:introduction}
Reinforcement learning (RL) is a cornerstone of modern artificial intelligence, enabling breakthroughs in complex autonomous systems and, more recently, the post-training and alignment of Large Language Models (LLMs) \citep{mnih2015human, silver2016mastering, silver2018general, ouyang2022training, rafailov2023direct}. However, these successes often rely on rich feedback, such as expert demonstrations in robotics or preference labels in language model alignment \citep{christiano2017deep, hester2018deep,brohan2022rt}. In the settings we study---where rewards are sparse or the goal is to surpass human performance---agents must go beyond imitation and autonomously discover novel behavior.

This requires \emph{active exploration}: an algorithmic process that deliberately collects informative experience \citep{ shyam2019model, sekar2020planning,eysenbach2018diversityneedlearningskills}. 
In the hardest cases, this amounts to ``finding a needle in a haystack,'' where a single rare trajectory is the key that unlocks learning. In structured settings like two-player games, self-play facilitates exploration \citep{silver2016mastering}. The adversarial dynamic ensures that novel strategies are always within reach of each agent's current policy. Outside self-play, driving exploration is significantly harder. Even in simple 2D environments, current methods often fail to explore efficiently without domain engineering; \montezuma and \pitfall remain standard benchmarks that stress-test systematic exploration \citep{ecoffet2019go, badia2020agent57b, badia2020never,guo2022byol}. These persistent failures highlight that systematic exploration in sparse-reward settings remains poorly understood. Unlocking efficient exploration would have impact well beyond games, enabling autonomous discovery in various scientific domains.

Historically, the primary approach to exploration relies on training agents with \emph{intrinsic motivation} \citep{schmidhuber1991curious, oudeyer2007intrinsic, barto2012intrinsic,houthooft2016vime, stadie2015incentivizing}. In this paradigm, the agent is trained to maximize a combined objective: the task reward plus an auxiliary signal designed so that maximizing it encourages the agent to visit novel states. In high-dimensional spaces, this auxiliary signal, commonly referred to as \emph{intrinsic reward} or \emph{bonus}, typically consists of a proxy for ``visitation frequency,'' such as the error of Random Network Distillation (\rnd) or the prediction error of a dynamics model \citep{burda2018exploration, pathak2017curiosity, pathak2019self}. While accurate uncertainty estimation is crucial, we argue that driving exploration by maximizing these signals via RL incurs unnecessary complexity; iteratively updating a policy to reach novel states is inherently sample-inefficient, as it requires constantly tracking a non-stationary intrinsic reward signal.

\paragraph{Contributions}
\arxiv{
\begin{itemize}[nosep, leftmargin=*]
\item We propose a new paradigm for autonomous exploration in sparse-reward settings that bypasses intrinsic-reward policy optimization. Our framework pairs a tree-search strategy, inspired by the Go-With-The-Winner (\gwtw) algorithm \citep{aldous1994go}, with epistemic uncertainty (a measure of the agent's uncertainty about a state given limited exposure to it) \citep{depeweg2018decomposition} to drive exploration; the approach is modular and agnostic to the specific uncertainty metric. The method relies on the ability to \emph{reset} the environment to previously visited states in order to redistribute computational effort toward the most promising frontiers. By replacing intrinsic-reward policy optimization with uncertainty-guided tree search, our method discovers high-reward trajectories using an order of magnitude fewer environment interactions than standard intrinsic motivation baselines on hard-exploration benchmarks.
\item We show that supervised backward learning algorithms (e.g.,~\cite{salimans2018learning}) can distill the generated trajectories into deployable, high-scoring policies. In particular, we achieve state-of-the-art results by a wide margin on \montezuma, \pitfall, and \venture when not relying on domain-specific knowledge.
\item We demonstrate the generality of our framework by applying it to high-dimensional continuous action spaces, solving the MuJoCo Adroit dexterous manipulation and AntMaze navigation tasks from pixel observations in a \emph{sparse-reward} setting without relying on expert demonstrations or offline datasets. To the best of our knowledge, this has not been previously achieved for the Adroit tasks. In both cases, the final distilled policies operate directly from image observations, without accessing privileged state information.
\end{itemize}}
\neurips{We propose a new paradigm for autonomous exploration in sparse-reward settings that bypasses intrinsic-reward policy optimization. Our framework pairs a tree-search strategy, inspired by the Go-With-The-Winner (\gwtw) algorithm \citep{aldous1994go}, with epistemic uncertainty (a measure of the agent's uncertainty about a state given limited exposure to it) \citep{depeweg2018decomposition} to drive exploration; the approach is modular and agnostic to the specific uncertainty metric. The method relies on the ability to \emph{reset} the environment to previously visited states in order to redistribute computational effort toward the most promising frontiers. By replacing intrinsic-reward policy optimization with uncertainty-guided tree search, our method discovers high-reward trajectories using an order of magnitude fewer environment interactions than standard intrinsic motivation baselines on hard-exploration benchmarks.

    We show that supervised backward learning algorithms (e.g.,~\cite{salimans2018learning}) can distill the generated trajectories into deployable, high-scoring policies. In particular, we achieve state-of-the-art results by a wide margin on \montezuma, \pitfall, and \venture when not relying on domain-specific knowledge. We demonstrate the generality of our framework by applying it to high-dimensional continuous action spaces, solving the MuJoCo Adroit dexterous manipulation and AntMaze navigation tasks from pixel observations in a \emph{sparse-reward} setting without relying on expert demonstrations or offline datasets. To the best of our knowledge, this has not been previously achieved for the Adroit tasks. In both cases, the final distilled policies operate directly from image observations, without accessing privileged state information.}

\paragraph{Outline} 
The remainder of this paper is organized as follows. \cref{sec:relatedwork} presents the related work and contextualizes our approach within the literature on exploration. \cref{sec:preliminaries} formalizes the problem setting and introduces core components, including the uncertainty estimator and the reset mechanism. \cref{sec:method} details our uncertainty-driven tree-search algorithm for exploration, \textsc{Go-With-Uncertainty} (\gowu). Finally, \cref{sec:experiments} evaluates the method on hard-exploration Atari games and challenging MuJoCo continuous-control tasks, showing efficient exploration and successful policy distillation across all environments.

\section{Related Work} 
\label{sec:relatedwork}
Efficient exploration remains a central challenge in reinforcement learning. Below we situate our approach within the broader landscape of exploration methods.

\paragraph{Intrinsic motivation and latent predictive models}
A dominant approach to exploration in deep RL is to train a policy to maximize an intrinsic motivation signal that encourages the agent to visit novel states \citep{stadie2015incentivizing,bellemare2016unifying,pathak2019self,badia2020agent57b}. In high-dimensional settings, these signals are often defined using proxies for novelty such as prediction error \citep{pathak2017curiosity}, pseudo-counts \citep{bellemare2016unifying}, Random Network Distillation \citep{burda2018exploration}, or information gain over a learned dynamics model \citep{houthooft2016vime}. Recent methods such as BYOL-Explore \citep{guo2022byol} continue this line by defining the intrinsic signal through the error in predicting future latent representations, an elegant way to measure novelty in high-dimensional observations. BYOL-Hindsight \citep{jarrett2023curiosityhindsightintrinsicexploration} extends this approach to better handle stochastic environments. These methods are among the strongest intrinsic-motivation baselines on hard-exploration benchmarks, yet even they remain limited on the hardest sparse-reward problems (such as \montezuma and \pitfall), especially once stochasticity is introduced. This supports our view that maximizing intrinsic motivation signals through standard policy optimization may be fundamentally inefficient for hard exploration.

\paragraph{Separating exploration from exploitation}
Another line of work separates exploration and downstream task optimization into distinct phases. In the empirical literature, this perspective appears in reward-free or unsupervised RL setups such as URLB \citep{laskin2021urlb}, where an agent first performs task-agnostic pre-training and is only later adapted to downstream rewards. Methods such as APT \citep{liu2021behavior} and Plan2Explore \citep{sekar2020planning} allocate this pre-training phase to broad state discovery using intrinsic objectives such as state coverage or model uncertainty, while skill-discovery methods such as variational intrinsic control \citep{mohamed2015variational} and DIAYN \citep{eysenbach2018diversityneedlearningskills} learn diverse behaviors before task rewards are introduced. This perspective also appears in the theoretical literature. A canonical early example is E$^3$ \citep{kearns2002near}, which formalizes an explicit explore-or-exploit strategy. The reward-free exploration framework of \citet{jin2020reward} goes further by showing that a pure exploration phase can suffice for efficient downstream planning in tabular MDPs. Subsequent work studies related ideas in rich-observation settings with discrete \citep{misra2020kinematic,mhammedi2023representation} and low-rank transition structure \citep{mhammedi2023efficient}.

\paragraph{Exploration without policy optimization} 
While the approaches above separate exploration from exploitation, the exploration phase itself still relies on policy optimization to maximize an intrinsic objective. Our approach departs from this by decoupling exploration from policy optimization entirely. To our knowledge, among methods that go beyond tabular settings, the only other approach that does this is Go-Explore \citep{ecoffet2019go,ecoffet2021first}. Both methods treat exploration as a search problem rather than as the optimization of an intrinsic reward, but the search mechanisms are fundamentally different; instead of maintaining an archive over discretized cells and repeatedly returning to archived states, we use a particle-based tree-search procedure guided by an epistemic uncertainty signal. Crucially, our method does not rely on hand-designed observation discretization or deterministic dynamics (Latent Go-Explore bypasses both but has other limitations; see \cref{app:additional_related}), making it more naturally suited to high-dimensional and stochastic environments. Despite these differences, both approaches share a common role: generating successful trajectories that can later be distilled into deployable policies through backward learning \citep{salimans2018learning,ecoffet2019go}.

\paragraph{Leveraging resets as a computational primitive}
Both \goexplore and our approach rely on a shared computational primitive: the ability to reset the environment to a previously visited state. This capability is also central to search-based methods such as Monte Carlo Tree Search (MCTS), which repeatedly branch from simulator states during planning \citep{kocsis2006bandit,coulom2006efficient}. While arbitrary resets are often impractical on physical systems \citep{eysenbach2017leavetracelearningreset,gupta2021resetfreereinforcementlearningmultitask}, they are readily available in simulation, where much of modern RL training already takes place \citep{zhao2020sim,navarro2025reality}. Recent theoretical work shows that local simulator access enables efficient learning in settings where standard online RL is provably inefficient \citep{li2021sample,yin2023sampleefficientdeepreinforcement,mhammedi2024power}. Our work treats resets as a training-time tool for exploration: they allow the algorithm to repeatedly redirect computation toward promising frontier states, without requiring simulator access at inference time.

\paragraph{Relationship to MCTS}
Although \gowu and MCTS both build search trees using simulator access, they differ substantially in both purpose and mechanism. MCTS is a planning algorithm that learns a value function during training and uses it at inference time to guide a search tree for action selection \citep{kocsis2006bandit,coulom2006efficient}. \gowu serves a fundamentally different purpose: it is a purely training-time procedure that discovers successful trajectories but produces no policy or value function. The actions taken during search are discarded; only the discovered state sequences are retained, which a second phase then uses to learn a policy. Beyond this difference in purpose, the search procedures themselves are fundamentally different. \gowu is a population-based algorithm inspired by the Go-With-The-Winner principle \citep{aldous1994go}: a population of particles evolves concurrently through the state space, with explicit cloning, pruning, and winner-selection operations that have no analogue in MCTS (see \cref{app:additional_related} for further differences).

\paragraph{Evolutionary and population-based methods} \gowu's population-based nature also connects it to evolutionary methods, which drive exploration by evolving diverse populations of policies. However, unlike \gowu, which searches directly in state space, these methods operate in the parameter space of a policy, generating diversity by perturbing model parameters (see \cref{app:additional_related} for more details).

\section{Preliminaries}
\label{sec:preliminaries}
We consider the problem of exploration in a stochastic episodic Markov Decision Process (MDP) defined by the tuple $\mathcal{M} = (\mathcal{X}, \mathcal{A}, P, R)$. Here, $\mathcal{X}$ and $\mathcal{A}$ are potentially large or infinite state and action spaces, respectively; $P: \mathcal{X} \times \mathcal{A} \to \Delta(\mathcal{X})$ is the transition function, and $R: \mathcal{X} \times \mathcal{A} \to \mathbb{R}$ is a deterministic reward function. The rest of this section introduces the building blocks of our approach.

\paragraph{Survival and failure}
Episodic MDPs define termination conditions that mark the end of a trajectory. For instance, in games like Atari, an episode typically ends after \emph{all} lives are lost. Our framework allows for distinguishing between this standard termination and a stricter notion of \emph{failure} we use for early episode truncation. In the Atari context, for example, we may treat the loss of a single life as a failure event; similarly, in continuous control, a state from which further progress is impossible (e.g., a quadruped flipping upside down in the AntMaze environment) can be treated as a failure, even if the environment does not formally terminate the episode. We demonstrate that using this flexibility can significantly improve exploration efficiency.

To formalize this, we partition the state space into viable states (\textsc{Alive}) and failure states (\textsc{Dead}). We further define \textsc{Doom} states, $\mathcal{X}_{\textsc{Doom}} \subset \textsc{Alive}$, as viable states from which failure is inevitable regardless of the policy. Our approach is designed for settings where progress—accumulating rewards or task completion—remains feasible from any non-\textsc{Doom} viable state.

\paragraph{Go-With-The-Winner (GWTW)} To discover high-reward trajectories efficiently, our method builds on the Go-With-The-Winner (\gwtw) principle \citep{aldous1994go}.
Originally proposed for randomized search and sampling, \gwtw is especially effective at finding deep nodes in imbalanced trees---a task where naive approaches like Depth-First or Breadth-First search can be computationally inefficient (see \cref{app:gwtw_examples} for a concrete example). The algorithm operates by maintaining a population of $N$ particles starting at the root. At each step, every particle advances one level by following a random edge. Crucially, particles that reach a leaf are pruned and replaced by clones of particles that reached non-leaf nodes, designated as ``winners.'' Theoretical guarantees for reaching a target depth in \gwtw worsen gracefully with a measure of tree imbalance \citep{aldous1994go}. Our approach builds on this principle by viewing exploration in an MDP as a search for ``deep'' nodes, where depth is characterized by high cumulative reward and epistemic uncertainty.

To adapt \gwtw for exploration, we map ``leaves'' to \textsc{Dead} states and redefine ``winners'' as states combining high accumulated reward and (epistemic) uncertainty.

\paragraph{Uncertainty oracle} Our winner criterion depends on a measure of epistemic uncertainty---the lack of knowledge resulting from limited data, as opposed to the inherent stochasticity of the environment \citep{osband2018randomized,kendall2017uncertainties,burda2018exploration}. We assume access to an uncertainty estimator, $U$, that assigns a scalar score to each state and updates online: as a state is visited more, its uncertainty decreases. The framework is agnostic to the choice of $U$; in our experiments, we use \rnd prediction error as a proxy \citep{burda2018exploration}.

\paragraph{Reset oracle} To enable cloning (as in \gwtw), we require overwriting one particle's state with another's via a reset primitive (see \cref{app:checkpointing} for implementation details).

\paragraph{Backward algorithm} 
Our new exploration algorithm does not produce a policy; it produces trajectories---sequences of environment checkpoints recording states visited by the most successful particles. These trajectories can be used in a second phase to learn deployable policies. Although the trajectories may contain redundancies or suboptimal loops, they serve as self-generated demonstrations for backward learning algorithms \citep{salimans2018learning,ecoffet2019go}. This class of algorithms begins training near the end of a demonstration and progressively moves the starting state backward as the agent masters each suffix segment, creating a natural curriculum that significantly simplifies the reinforcement learning problem. In our experiments, we use this framework to obtain high-scoring policies across all environments (see \cref{app:backward} for full implementation details).

\section{Method: Uncertainty-Guided Exploration}
\label{sec:method}
Building on the primitives in \cref{sec:preliminaries}, we propose \textsc{Go-With-Uncertainty} (\gowu), a tree-search method decoupling exploration from policy optimization. Our approach adapts the \gwtw principle by maintaining a population of particles that explore the state space in parallel. At fixed intervals, the algorithm identifies ``winners''—those in states with high cumulative reward and epistemic uncertainty. Using the reset oracle, it redistributes computational effort by cloning winners and removing \dead particles.

 We first describe the state-lineage tree, which tracks particle history and enables population management. We then detail particle evolution and the winner selection scoring function, referencing the pseudo-algorithms \cref{alg:population_iteration,alg:parallel_evolution} along the way.
 
\iftoggle{neurips}{}{\begin{figure}[t]
    \centering
    \includegraphics[width=\linewidth, trim=0pt 90pt 0pt 20pt, clip]{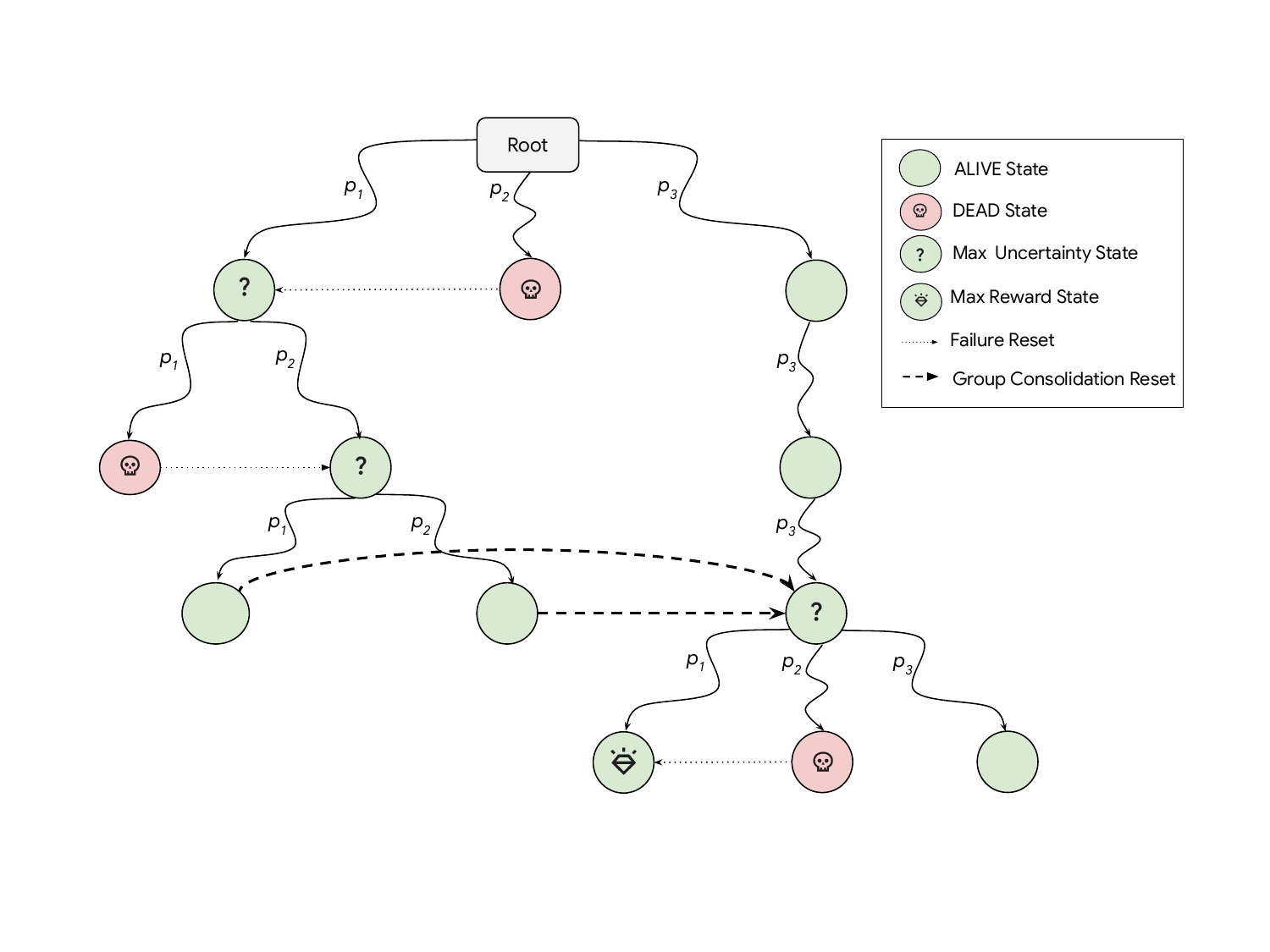}
    \caption{Schematic overview of \gowu with a single group of particles. The algorithm maintains a population of particles ($p_1, p_2, p_3$) that explore the state space via multi-step rollouts. During an outer step, if a particle reaches a \dead state (e.g., $p_2$ on level one), it is pruned and its state is reset via a \emph{Reset} to the winner---the \alive node maximizing accumulated reward, with epistemic uncertainty used as a tie-breaker. After $K$ outer steps, a \emph{Group Consolidation Reset} syncs all particles, including \alive particles, to the current winner.}
    \label{fig:gowu_schematic}
\end{figure}
}

\subsection{The State Lineage Tree}
\label{sec:lineage_tree} 
To manage populations, \gowu maintains a global \emph{state-lineage tree}. Each node in this tree represents a valid, checkpointed state of the environment. Formally, a tree node $v$ is a tuple $(s_v, R_v, u_v)$, where $s_v$ is the compact state representation (e.g., simulator snapshot); $R_v$ is the cumulative extrinsic reward achieved by the agent up to state $s_v$; and $u_v$ is a pointer to the \parent node. 

This structure allows the algorithm to track the history of every particle. To manage memory, nodes are pruned when they are no longer occupied by an active particle or referenced by any descendant. This ensures that the storage requirements scale only with the set of active lineages. 

\subsection{Particle Evolution}
\label{sec:particleevol}
We now describe the search procedure, which implements our adaptation of \gwtw to drive exploration using a group of $N$ particles (e.g., $N=32$ for some of our experiments). The search proceeds in discrete iterations. \cref{alg:population_iteration} details the execution of a single iteration, which consists of a sequence of $K$ \outersteps, followed by a synchronization phase, where $K$ is sampled uniformly from $[K_{\min}, K_{\max}]$ at the start of each iteration. 

\paragraph{Evolutionary rollout} 
During an outer step, every \alive particle interacts with its own instance of the environment for a random number of simulation steps (\innersteps) sampled independently for each particle from the range $[T_{\min}, T_{\max}]$ (see \cref{line:sample}). Our framework is agnostic to the mechanism each particle uses to select actions; in practice, we employ random policies as described in \cref{sec:implementation}. During the \innersteps, the uncertainty estimator $U$ may update based on newly collected observations. If a particle receives a positive reward, the rollout is terminated immediately (\cref{line:rewardbreak}); this secures the reward for the subsequent winner selection, as further steps may lead to a \dead state. If a particle completes the rollout without entering a \dead state, the algorithm creates a new child node in the lineage tree (\cref{line:addchild}). The rollout length range controls a key trade-off: overly short rollouts limit state diversity, while overly long ones increase the frequency of \dead states. (In our experiments, $T_{\min}$ and $T_{\max}$ are set between $3$ and $20$.)

\paragraph{Failure recovery via rollback}
If the entire population enters a \textsc{Dead} state during a rollout, the algorithm performs a collective \rollback. Every particle traverses the lineage tree via parent pointers $u_v$ to a randomly selected ancestor between $k_{\min}$ and $k_{\max}$ generations prior (\cref{line:rollback}). This restores the particles to valid antecedent states, allowing exploration of alternative trajectories around, for example, obstacles. A sufficiently large $k_{\max}$ allows particles to escape \emph{doom states} (e.g., irreversible falls), while a low $k_{\min}$ maintains proximity to difficult obstacles, enabling alternative trajectory attempts ($k_{\min}, k_{\max} \in [1, 20]$ in our experiments).

\paragraph{Winner selection and particle redistribution}
\label{sec:reassignment}
When at least one particle survives the \innersteps rollout, the algorithm redistributes computational effort by identifying ``winners.'' \begin{enumerate}[nosep, leftmargin=*]\item \emph{Winner selection:} The algorithm identifies the \alive particles with the maximum accumulated reward. Among these, it selects the particle with the highest uncertainty $U(s)$ (\cref{line:winner}). This prioritizes high-value, under-explored states; in environments with sparse rewards (like the ones in our experiments), uncertainty is often the primary criterion for selection. \label{item:winnerselection}\item \emph{Pruning:} All \dead particles are immediately reset to the winner's state (\cref{line:prune}). This prunes failed branches and redirects those resources to the current frontier. Non-winner \alive particles continue independent trajectories to maintain local diversity.\end{enumerate}

\paragraph{Group consolidation}
At the conclusion of the $K$ \outersteps, the algorithm performs an additional synchronization to consolidate progress. We identify a single group winner based on accumulated reward and uncertainty as in \cref{item:winnerselection}. Unlike the mid-iteration redistribution, where only \dead particles are reset, this step collapses the entire group to the winner's state (\cref{line:single}), ensuring the next iteration begins from the most promising frontier discovered so far.

\subsection{Parallel Groups and Population Synchronization}
\label{sec:parallel}
To diversify the search and accelerate discovery, \gowu runs $M$ groups in parallel; see \cref{alg:parallel_evolution} (e.g., $M=4$ for our Atari experiments). While these groups may execute rollouts independently, our framework allows the groups to share a centralized uncertainty estimator. This enables implicit coordination: updates from one group discourage others from revisiting the same states, pushing the collective population toward globally novel regions. To further prevent groups from stagnating in explored regions, the algorithm employs a global synchronization mechanism at the start of each iteration. We identify a single ``global winner'' across all $M$ groups based on the reward-uncertainty criterion in \cref{item:winnerselection} (\cref{line:identify}). The algorithm then compares each group's maximum reward to the global winner's. If a group's best reward is strictly lower, all its particles are reset to the global winner (\cref{line:resetgroup}). If equal, the group continues its search independently. This mechanism balances diversity with efficiency: it allows competitive groups to explore unique trajectories while propagating breakthrough discoveries across the entire population.

\iftoggle{neurips}{}{\begin{algorithm}[t]
\caption{\evolvegroup}
\label{alg:population_iteration}
\begin{algorithmic}[1]
\item[] \textbf{input:} Population $P=\{p_1, \dots, p_N\}$, lineage tree $\mathcal{T}$, uncertainty $U$
\item[] \textbf{params:} \outersteps, \innersteps, and \rollback ranges: $[K_{\min},K_{\max}]$, $[T_{\min},T_{\max}]$, $[k_{\min},k_{\max}]$
\vspace{0.2cm}
\STATE Sample \outersteps $K \sim \text{Unif}(K_{\min}, K_{\max})$
\FOR{$k=1, \dots, K$}
    \FOR{each particle $p_i \in P$}  
        \STATE Sample \innersteps $T_i \sim \text{Unif}(T_{\min}, T_{\max})$   \mline{line:sample}
        \STATE  \algcommentbiglight{Rollout particle and update uncertainty}
        \FOR{$t = 1, \dots, T_i$} \label{line:step}
            \STATE $U \gets p_i.\step(1, U)$ \hfill \algcommentlight{Execute one step}
            \STATE \textbf{if} {$p_i$ is \dead or $p_i.R$ increased} \textbf{then} \textbf{break} \hfill\algcommentlight{Halting on reward does not kill the particle} \mline{line:rewardbreak}
        \ENDFOR
        \IF{$p_i$ is \textsc{alive}}
            \STATE $\mathcal{T}.\addchild(p_i)$\hfill \algcommentlight{Expand tree for surviving particles} \mline{line:addchild}
            \ENDIF
    \ENDFOR

    \IF{All $p_i \in P$ are \textsc{Dead}} \mline{line:alldead}
        \STATE  \algcommentbiglight{Failure recovery via rollback}
        \STATE $v_{i,\anc} \gets p_i.\getancestor(k_{\min}, k_{\max})$, $\forall i$ \hfill \algcommentlight{Pick $k_i$th ancestor of $p_i$ with $k_i\sim \text{Unif}(k_{\min},k_{\max})$, $\forall i$} 
        \STATE $P \gets \{ \reset(p_i, v_{i,\anc}) \mid \forall i \}$ \hfill \algcommentlight{Restore all particles to ancestor checkpoints} \mline{line:rollback}
    \ELSE
        \STATE  \algcommentbiglight{Uncertainty-aware redistribution}
        \STATE $p_{\winner} \gets \lexargmax_{p \in P, p.\textsc{alive}} (p.R, U(p))$ \hfill \algcommentlight{Best alive: max reward, break ties by uncertainty} \mline{line:winner}
        \FOR{each $p_i \in P$}
            \IF{$p_i$ is \textsc{Dead}}
                \STATE $\reset(p_i, p_{\winner})$ \hfill \algcommentlight{Prune \& respawn; $p_i$ becomes \alive after the reset} \mline{line:prune}
            \ENDIF
        \ENDFOR
    \ENDIF
\ENDFOR
\STATE  \algcommentbiglight{Group consolidation (lexicographic ranking)}
\STATE $p_{\winner} \gets \lexargmax_{p \in P} (p.R, U(p))$ \mline{line:groupwinner}
\STATE  \mline{line:single}$P \gets \{ \reset(p_j, p_{\winner}) \mid \forall j \}$ \hfill \algcommentlight{Reset all particles to winner's checkpoint}
\STATE \textbf{return} $P,\mathcal{T}, U, p_{\winner}$
\end{algorithmic}
\end{algorithm}

\begin{algorithm}[t]
\caption{\gowu: \textsc{Go-With-Uncertainty}}
\label{alg:parallel_evolution}
\begin{algorithmic}[1]  
\item[] \textbf{input:} Env.~$\mathcal{E}$, iterations $N_{\text{iter}}$, number of groups $M$, and number of particles per group $N$
\vspace{0.2cm}
\STATE Initialize lineage tree $\mathcal{T}$ with root node $s_0$ (initial state)
\STATE Initialize groups $\{G_1, \dots, G_M\}$, each with $N$ particles, all starting at $s_0$
\STATE Initialize shared uncertainty estimator $U$
\STATE Initialize $p_{\winner}$ at state $s_0$ with cumulative reward $0$
\FOR{$i=1, \dots, N_{\text{iter}}$}
\FOR{each Group $G_m$ in parallel}
\STATE \algcommentbiglight{Global synchronization}
\STATE $p_{m} \gets \lexargmax_{p \in G_m} (p.R, U(p))$ \hfill \IF{$p_{m}.R < p_{\winner}.R$}
             \STATE Reset $G_m$ to the state of $p_{\winner}$ \hfill \algcommentlight{Restore all particles to global winner's checkpoint} \mline{line:resetgroup}
        \ENDIF
    \STATE \algcommentbiglight{Evolve group}
        \STATE $G_m, \mathcal{T}, U, p_{m} \gets\evolvegroup(G_m, \mathcal{T}, U)$\hfill \algcommentlight{\cref{alg:population_iteration}}
    \ENDFOR
\STATE \algcommentbiglight{Update global winner}
\STATE $p_{\winner} \gets \lexargmax_{p \in \{p_1, \dots, p_M, p_{\winner}\}} (p.R, U(p))$ \mline{line:identify}
\ENDFOR
\end{algorithmic}
\end{algorithm}
}

\subsection{Implementation Overview}
\label{sec:implementation}
We implement \gowu using a distributed coordinator-worker architecture that decouples environment simulation, population management, and uncertainty model training. Rollout workers stream observations into a shared replay buffer, from which a dedicated learner asynchronously trains the \rnd predictor \citep{burda2018exploration} that instantiates the uncertainty oracle $U$. Particles select actions using a simple random policy: each particle commits to a single random action held fixed for the duration of each rollout segment. In the discrete setting, a different random action is played with small probability at each step ($\varepsilon = 0.2$). Full implementation details and model architectures are provided in \cref{app:implementation}.

\section{Experiments}
\label{sec:experiments}
\label{sec:expsetup}
We evaluate \gowu on two families of hard-exploration benchmarks spanning both discrete and continuous action spaces: Atari games (\montezuma, \pitfall, and \venture) and MuJoCo continuous-control tasks (AntMaze and Adroit dexterous manipulation). Our experiments are designed to answer the following research questions:
\begin{itemize}[leftmargin=*, nosep]
    \item \emph{RQ1 (Exploration efficiency):} Can \gowu efficiently discover high-reward and task-completing trajectories across both discrete and continuous domains?
    \item \emph{RQ2 (Quality of exploratory data):} Can the discovered trajectories be distilled into deployable policies that achieve high scores (Atari) or high success rates (MuJoCo)?
\end{itemize}

\paragraph{Atari environments} We evaluate on \montezuma, \pitfall, and \venture---three of the most challenging exploration benchmarks in the Arcade Learning Environment (ALE) \citep{bellemare2013arcade,machado2018revisiting}. \montezuma requires long-horizon planning across multiple rooms, while \pitfall presents a distinct challenge due to its very sparse rewards and visually similar rooms. Notably, \pitfall remains largely unsolved without human demonstrations or domain knowledge \citep{ecoffet2021first} when using sticky actions. \venture adds a distinct challenge: pursuing enemies imposing strict time pressure, requiring the agent to explore certain rooms \emph{quickly}.
 
\iftoggle{neurips}{}{\begin{figure}[b]
    \centering
    \setlength{\tabcolsep}{4pt}
    \begin{tabular}{@{}ccc@{}}
        \includegraphics[height=3.4cm]{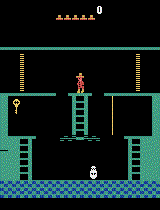} &
        \includegraphics[height=3.4cm]{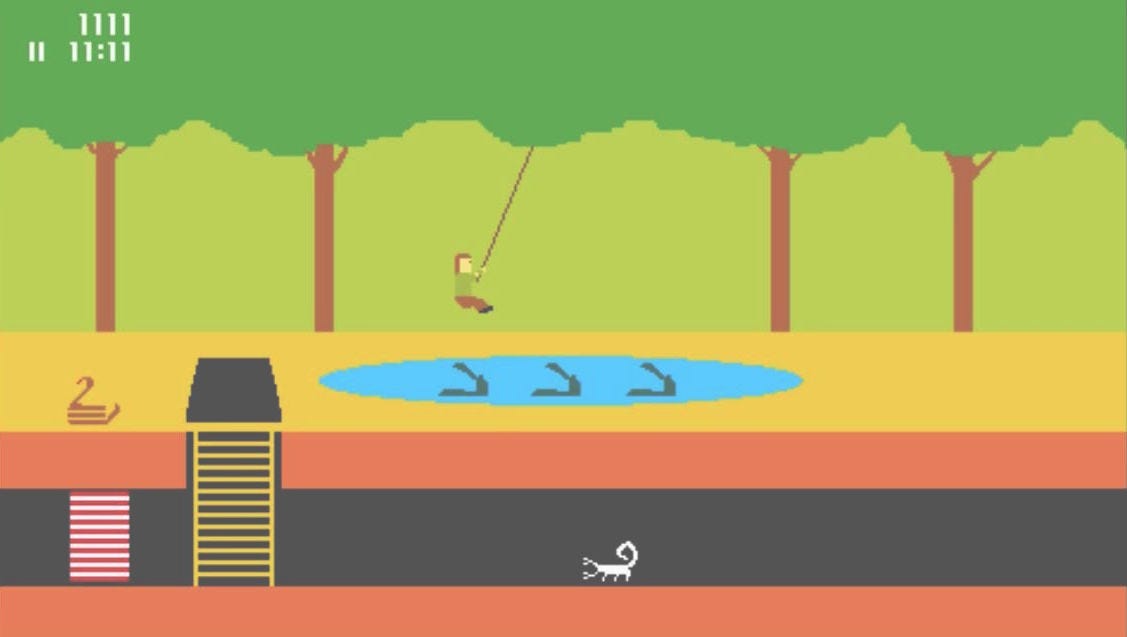} &
        \includegraphics[height=3.4cm]{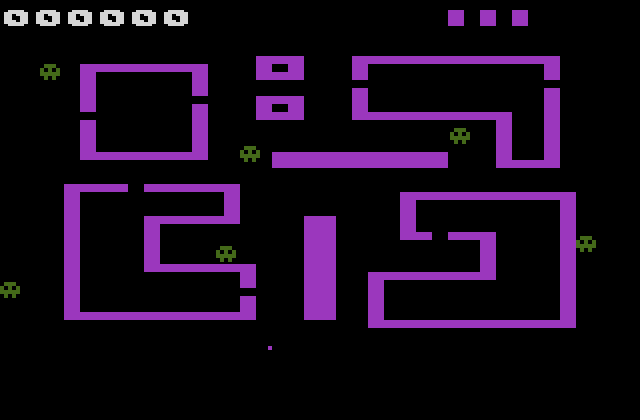} \\[4pt]
        {\small (a) \montezuma} & {\small (b) \pitfall} & {\small (c) \venture}
    \end{tabular}
    \caption{Fully rendered observations from the three hard-exploration Atari games used in our evaluation.}
    \label{fig:atari_games}
\end{figure}
}
\iftoggle{neurips}{}{\begin{figure}[t]
        \centering
        \begin{minipage}[b]{0.24\linewidth}
            \centering
            \includegraphics[width=\linewidth, trim=0pt 0pt 0pt 18pt, clip]{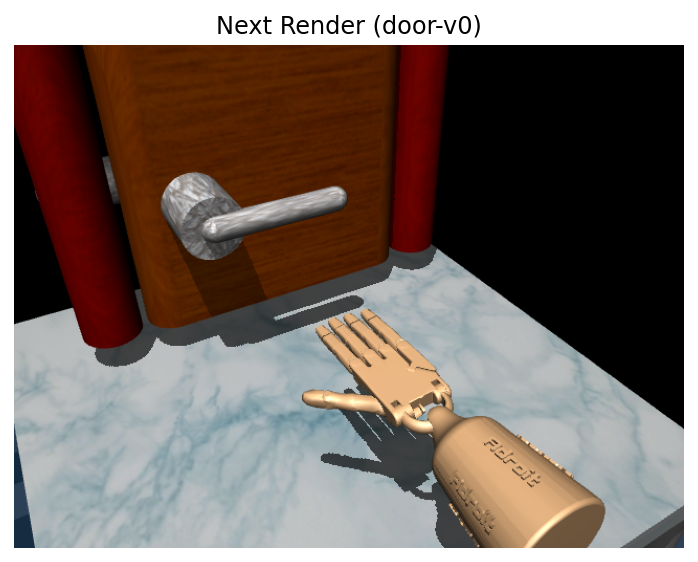}\\
            {\small (a) \texttt{door-v0}}
        \end{minipage}
        \hfill
        \begin{minipage}[b]{0.24\linewidth}
            \centering
            \includegraphics[width=\linewidth, trim=0pt 0pt 0pt 18pt, clip]{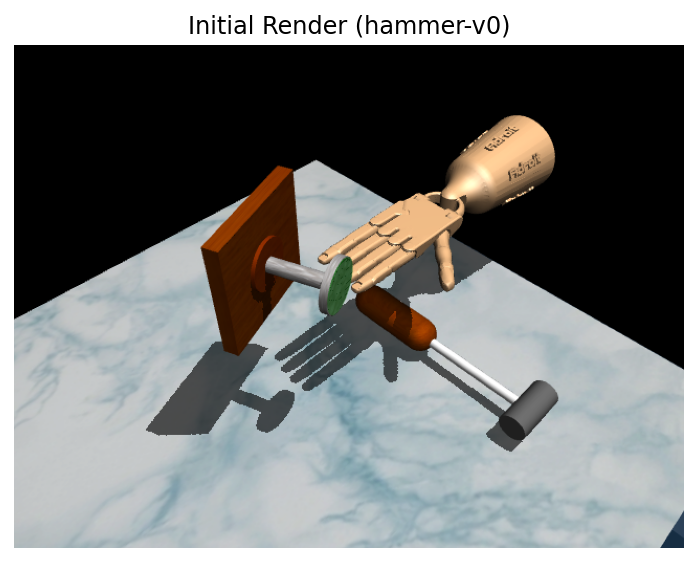}\\
            {\small (b) \texttt{hammer-v0}}
        \end{minipage}
        \hfill
        \begin{minipage}[b]{0.24\linewidth}
            \centering
            \includegraphics[width=\linewidth, trim=0pt 0pt 0pt 18pt, clip]{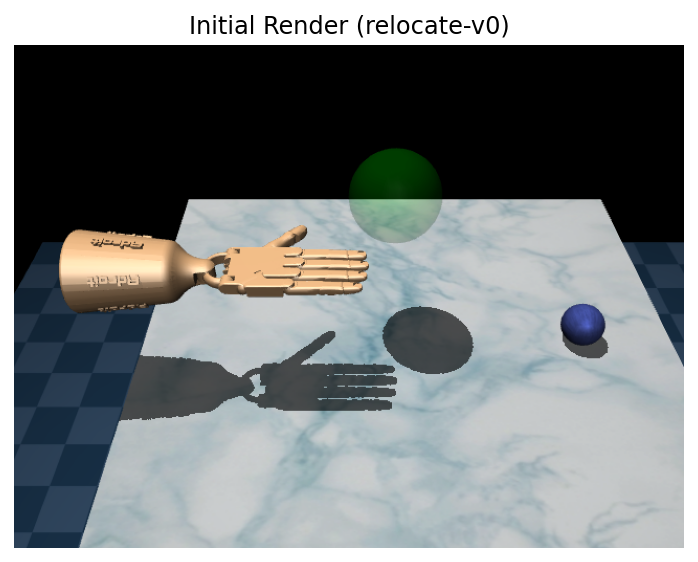}\\
            {\small (c) \texttt{relocate-v0}}
        \end{minipage}
        \hfill
        \begin{minipage}[b]{0.24\linewidth}
            \centering
            \includegraphics[width=\linewidth, trim=0pt 0pt 0pt 18pt, clip]{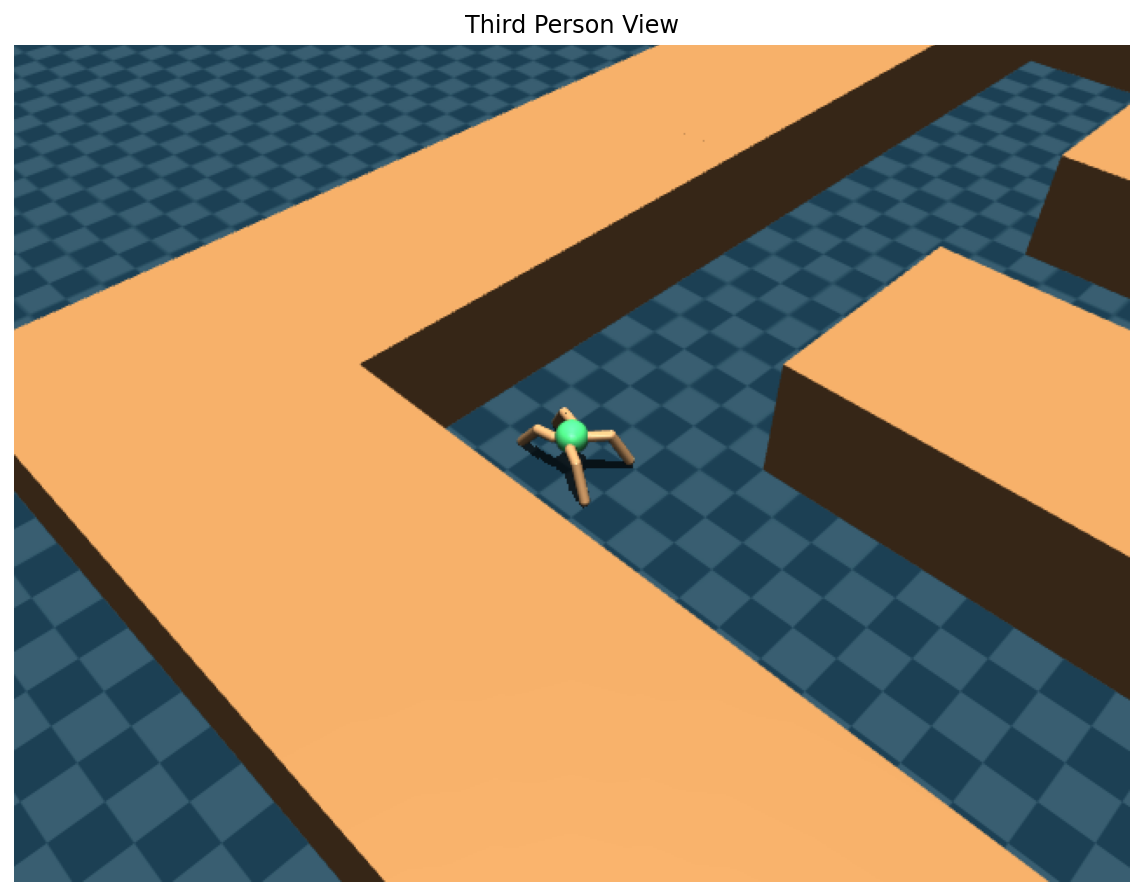}\\
            {\small (d) AntMaze}
        \end{minipage}
        \caption{MuJoCo continuous-control tasks used in our evaluation. (a--c)~Adroit dexterous manipulation tasks using the 24-DoF ShadowHand. (d)~AntMaze navigation task (top-down view).}
        \label{fig:mujoco_tasks}
    \end{figure}
}
\paragraph{MuJoCo environments} To evaluate \gowu on continuous action spaces, we consider two families of tasks that require deep exploration from visual observations:
\begin{itemize}[nosep, leftmargin=*]
\item \emph{AntMaze} (\texttt{antmaze-large-diverse-v0}): A quadruped ant must navigate a large maze to reach a goal position. The environment provides only a sparse terminal reward upon reaching the goal (the default setting), requiring sustained exploration through extended corridors and dead ends.
\item \emph{Adroit} (\texttt{door-v0}, \texttt{hammer-v0}, \texttt{relocate-v0}): Dexterous manipulation tasks using the 24-degree-of-freedom ShadowHand \citep{rajeswaran2017learning}. \texttt{door-v0} requires opening a door by its handle, \texttt{hammer-v0} requires picking up a hammer and driving a nail, and \texttt{relocate-v0} requires grasping a ball and moving it to a target location. The high dimensionality of the action space and the precision required for manipulation make these tasks extremely challenging. We enable the \texttt{sparse\_reward} flag, which replaces the default dense reward shaping with a single task-completion reward.
\end{itemize}
Although the MuJoCo dynamics are deterministic, task configurations are randomized across seeds, preventing the policy from memorizing a fixed action sequence. For AntMaze, the target location is slightly randomized around the top-right corner of the maze. For \texttt{door-v0}, the position of the door and handle are perturbed across episodes; similarly, the nail position varies in \texttt{hammer-v0}. For \texttt{relocate-v0}, the randomization is more substantial: both the ball and target locations change significantly between seeds. We aim to solve all MuJoCo tasks \emph{from images}: the final distilled policy receives only visual observations, not privileged state information.

\paragraph{Observations and environment configuration}
Observations are processed into grayscale, frame-stacked images. For Atari, we follow the standard preprocessing pipeline \citep{mnih2015human}, producing $84 \times 84 \times 4$ observations. For MuJoCo, we render task-specific camera views at $120 \times 120$ and stack 4 consecutive frames; for AntMaze, an additional global top-down view of the maze is included (during Phase~I, only this top-down view is used). All pixel values are normalized to $[0, 1]$. For the Atari environments, we use \emph{sticky actions} ($p = 0.25$), introducing aleatoric stochasticity \citep{machado2018revisiting}. For Atari and AntMaze, we use the standard action repeat of 4, where the selected action is applied 4 times; thus, each environment step consumes four game frames. Adroit tasks use no action repeat owing to the precision required for dexterous manipulation. Full details on observation processing are provided in \cref{app:obs_processing}. Finally, we disable the \pitfall timer during Phase~I to enable non-episodic exploration; the timer is re-enabled during Phase~II and evaluation (see \cref{sec:resets} for details).

\iftoggle{neurips}{}{\begin{figure}[t]
\centering
\newcommand{\obsimg}[1]{\raisebox{-0.5\height}{\includegraphics[width=0.14\linewidth, trim=0pt 0pt 0pt 18pt, clip]{figures/#1}}}
\begin{tabular}{@{}m{4em} cccc c@{}}
 & \textbf{Frame $-3$} & \textbf{Frame $-2$} & \textbf{Frame $-1$} & \textbf{Frame $0$} & \textbf{Frame top} \\[2pt]
\texttt{door}     & \obsimg{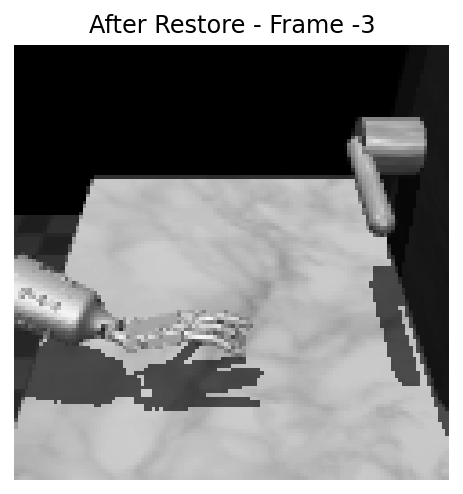} & \obsimg{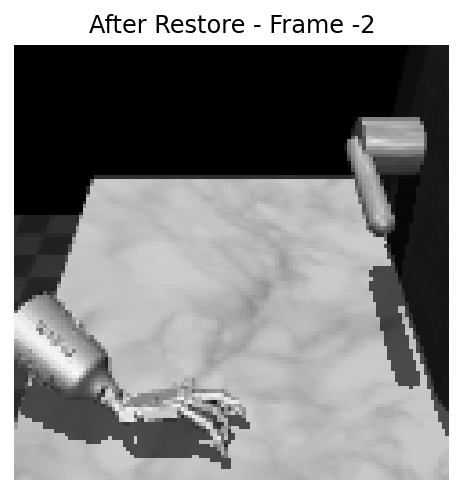} & \obsimg{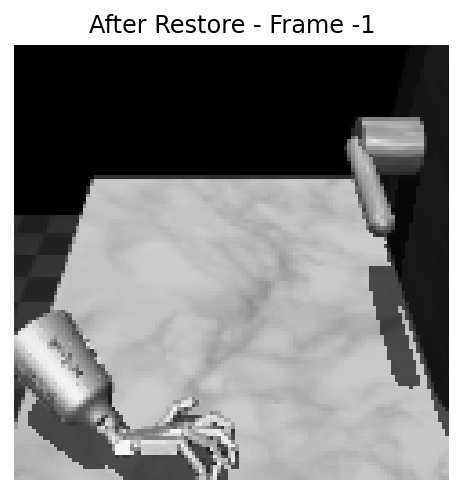} & \obsimg{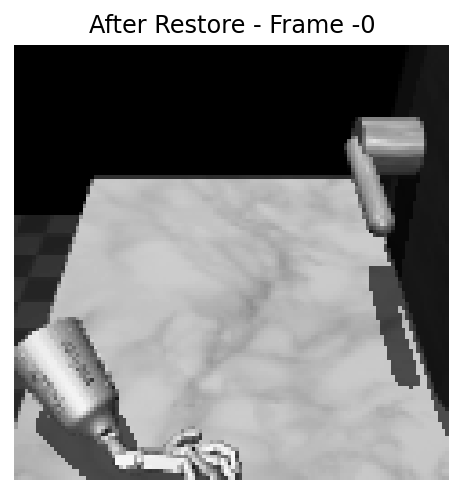} & \\[2pt]
\texttt{hammer}   & \obsimg{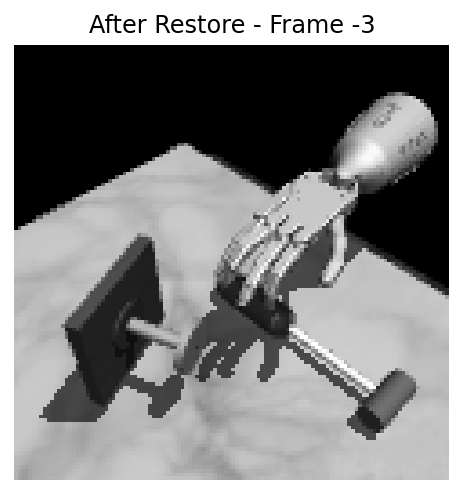} & \obsimg{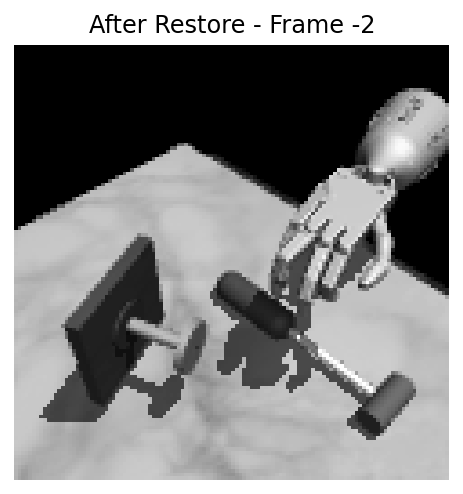} & \obsimg{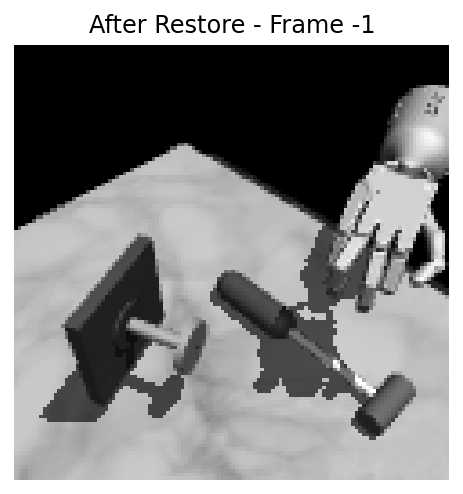} & \obsimg{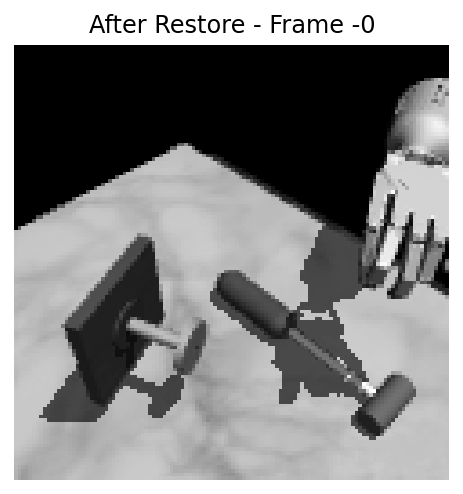} & \\[2pt]
\texttt{relocate} & \obsimg{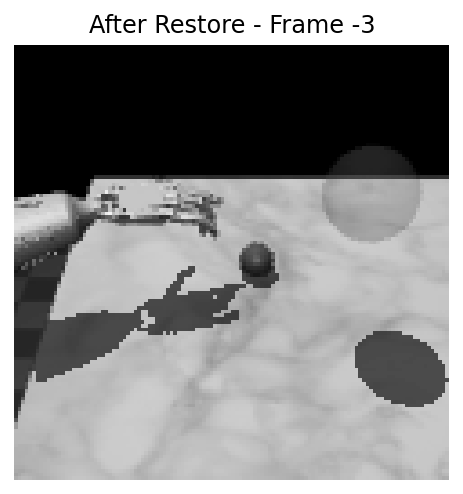} & \obsimg{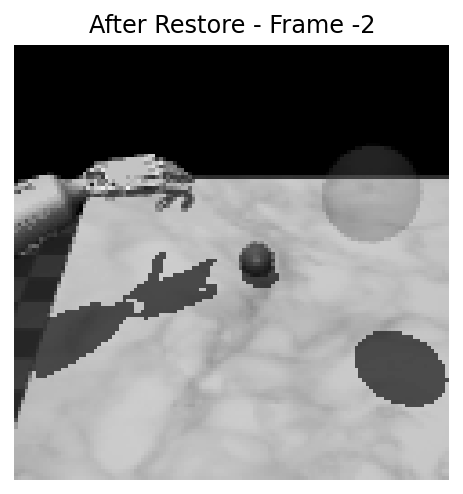} & \obsimg{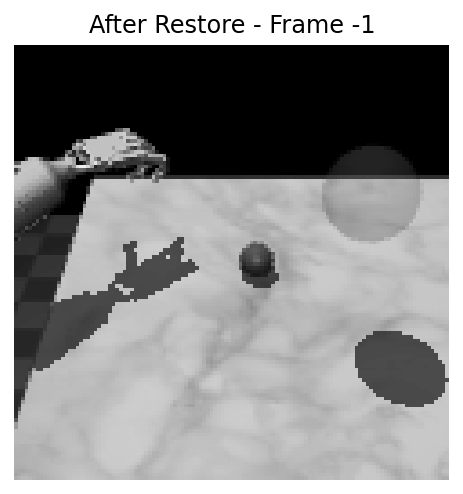} & \obsimg{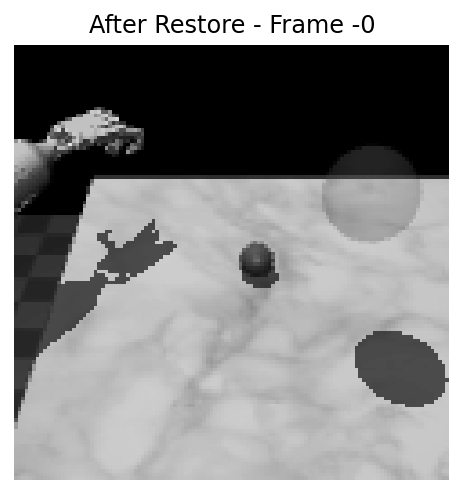} & \\[2pt]
\texttt{antmaze}  & \obsimg{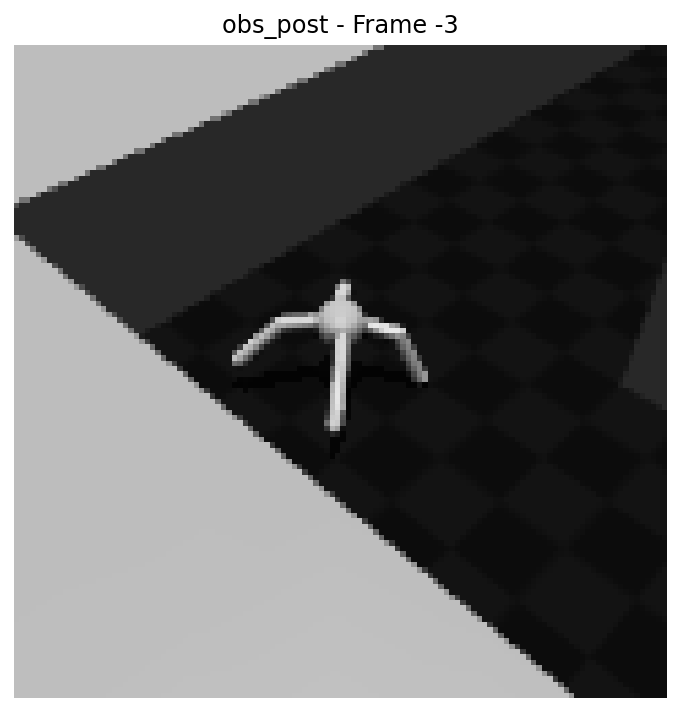} & \obsimg{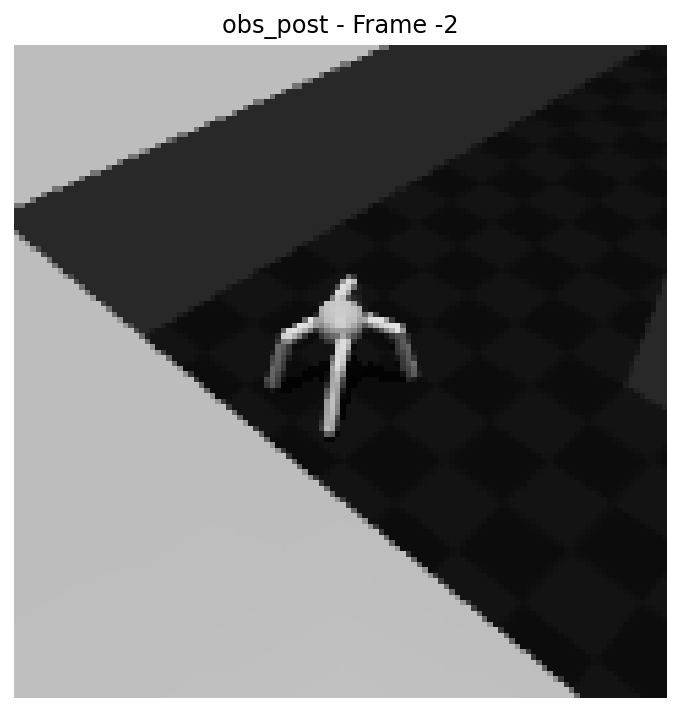} & \obsimg{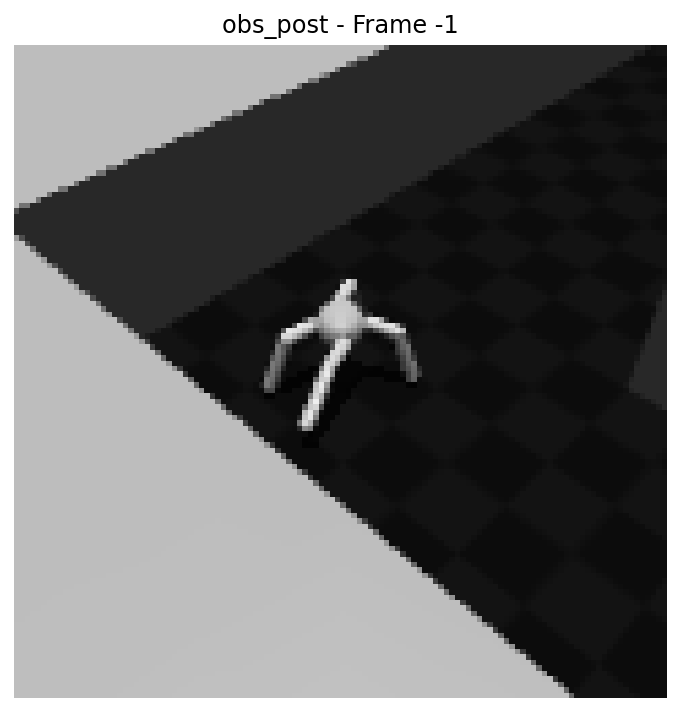} & \obsimg{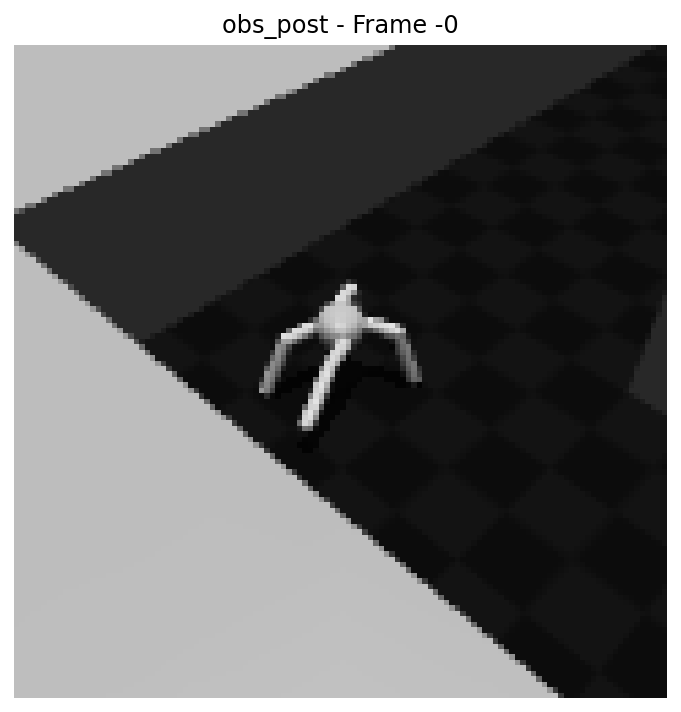} & \obsimg{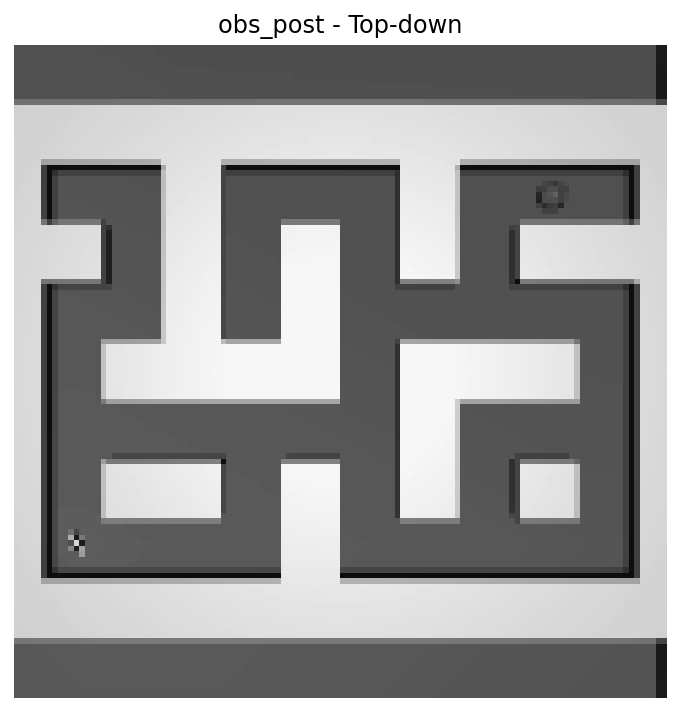} \\
\end{tabular}
\caption{Processed visual observations as seen by the agent for each MuJoCo task. Each column shows a frame in the observation stack, from the oldest (Frame~$-3$) to the most recent (Frame~$0$). For AntMaze, the ``Frame top'' column shows the global top-down view of the maze; during Phase~I (exploration), only this top-down view is used. See \cref{sec:expsetup} for an overview and \cref{app:obs_processing} for full details on the observation processing pipeline.}
\label{fig:obs_processing}
\end{figure}
}
 
\paragraph{Exploration signals (Phase~I)} During Phase~I, we define environment-specific \dead state criteria---and, for MuJoCo, additional intermediate rewards---to guide particle redistribution in \gowu.
\begin{itemize}[nosep, leftmargin=*]
\item \emph{Atari:} We treat a loss of life---detected via a discount factor of zero---as a \dead state. In \pitfall, we additionally treat any negative reward (e.g., from touching a rolling barrel) as \dead, since it signals an unrecoverable penalty. For \montezuma, we define an additional \dead state to prevent exploitation of a newly discovered environment bug (see \cref{app:middleroombug}).
\item \emph{AntMaze:} We define flipping over as a \dead state, since the ant becomes irrecoverably stuck. See \cref{app:mujoco_rewards} for details on how this condition is extracted.
\item \emph{Adroit:} We define a single intermediate reward based on contact with the target surface: a reward of $+1$ is assigned when the hand first contacts the relevant object---the door handle, the hammer, or the ball. If contact is subsequently lost, we mark the particle as \dead.
\end{itemize}
We emphasize that these signals are straightforwardly extracted from readily available state information and are far from the heavy reward shaping typically required by standard RL algorithms. 

\paragraph{\gowu parameters (Phase~I)} Population management parameters we use are listed in \cref{tab:hyperparams}.

\paragraph{Policy distillation (Phase~II)}
\label{sec:backwardimplement}
Following the exploration phase, we distill the discovered trajectories into deployable policies using only the original environment rewards (the additional exploration signals from Phase~I are not used). For each \gowu run, we extract the highest-reward trajectory from the lineage tree (\cref{sec:lineage_tree}) and use it as a demonstration for a backward learning curriculum \citep{salimans2018learning}. The agent is initialized near the end of the demonstration and trained using \textsc{PPO} \citep{schulman2017proximalpolicyoptimizationalgorithms} to maximize cumulative reward from that point forward; as it masters the task, the initialization is progressively moved backward toward the start of the episode. For Atari, we decompose each demonstration into 10 segments delineated by reward events, training on all segments simultaneously; this decomposition breaks the long-horizon problem into manageable sub-tasks. For MuJoCo, no segmentation is applied and the goal location is held fixed to the one used during exploration while the curriculum progresses backward along a given demonstration. Once the curriculum reaches the initial state, the goal (i.e., the randomized target location and object placement described above) is allowed to vary across episodes, training the policy to generalize; using 10 demonstrations per Phase~II run, each corresponding to a different goal configuration, provides initial diversity across goals. Throughout training, a background evaluator periodically fetches the latest policy weights and tracks the best-performing checkpoint. After training is complete, this checkpoint is evaluated on $500$ rollouts to produce the final policy score.

See \cref{app:backward} for additional details on the curriculum strategy, environment-specific configurations, and evaluation procedures.

\paragraph{Baselines} For Atari, we compare against \goexplore (no domain knowledge) \citep{ecoffet2021first}, which represents the current state of the art on hard-exploration games, and three intrinsic motivation methods: \textsc{RND} \citep{burda2018exploration}, \textsc{MEME} \citep{kapturowski2022human} (\textsc{Agent57}'s successor \citep{badia2020agent57b}), and \textsc{BYOL-Hindsight} \citep{jarrett2023curiosityhindsightintrinsicexploration}. The latter two are among the strongest intrinsic motivation approaches for hard-exploration games.

For the MuJoCo tasks, direct baselines are unavailable. Existing methods for Adroit and AntMaze either rely on dense reward shaping, use privileged state observations, or operate in simpler settings. To the best of our knowledge, no prior method solves the adroit tasks from pixel observations in the sparse-reward setting without expert demonstrations. We discuss this further in \cref{app:additional_related}.

\subsection{Phase I Results: Exploration Efficiency of \gowu}
\label{sec:results}
\begin{table}[ht]
\centering
\caption{Comparison of \gowu against baselines on hard-exploration Atari games. For \gowu, we report the mean ($\pm$ standard deviation) cumulative reward along the discovered demonstrations across 100 exploration seeds after at most 600M frames per seed, and the mean ($\pm$ standard deviation) policy score across 10 Phase~II runs (see \cref{tab:robustification_stats} for full statistics). All results use sticky actions ($p = 0.25$), except \goexplore, which uses a deterministic environment during exploration. Baseline numbers are taken from the respective papers. $^*$\textsc{BYOL-Hindsight} values are approximate, extracted visually from Figure 19 in the original paper.}
\label{tab:main_comparison}
\begin{sc}
\begin{small}
\resizebox{\columnwidth}{!}{\begin{tabular}{lcccccc}
\toprule
\multirow{2}{*}{\textbf{Game}} & \multicolumn{2}{c}{\textbf{Ours}} & \multirow{2}{*}{\textbf{\goexplore}} & \multirow{2}{*}{\textbf{RND}} & \multirow{2}{*}{\textbf{MEME}} & \multirow{2}{*}{\textbf{BYOL-Hind.}} \\
\cmidrule(lr){2-3}
 & Exploration (\gowu) & Robustification & & & & \\
\midrule
Montezuma & 98{,}249{ $\pm$18{,}102} & {\bf 196{,}312}{ $\pm$49{,}670}  & 43{,}791 & 8{,}152 & 9{,}429 & ${\sim}14{,}517^*$ \\
Pitfall   & 54{,}440{ $\pm$14{,}817} & {\bf 101{,}972}{ $\pm$2{,}226}  & 6{,}945 & $-3$ & 7{,}821 & ${\sim}16{,}211^*$ \\
Venture   & 5{,}012{ $\pm$2{,}678} & {\bf 6{,}436}{ $\pm$1{,}504} & 2{,}281 & 1{,}859 & 2{,}583 & ${\sim}2{,}328^*$ \\
\bottomrule
\end{tabular}}
\end{small}
\end{sc}
\end{table}

\begin{figure}[ht]
    \centering
    \resizebox{\columnwidth}{!}{\begin{tabular}{ccc}
        \includegraphics[width=0.31\linewidth, trim=0pt 0pt 0pt 0pt, clip]{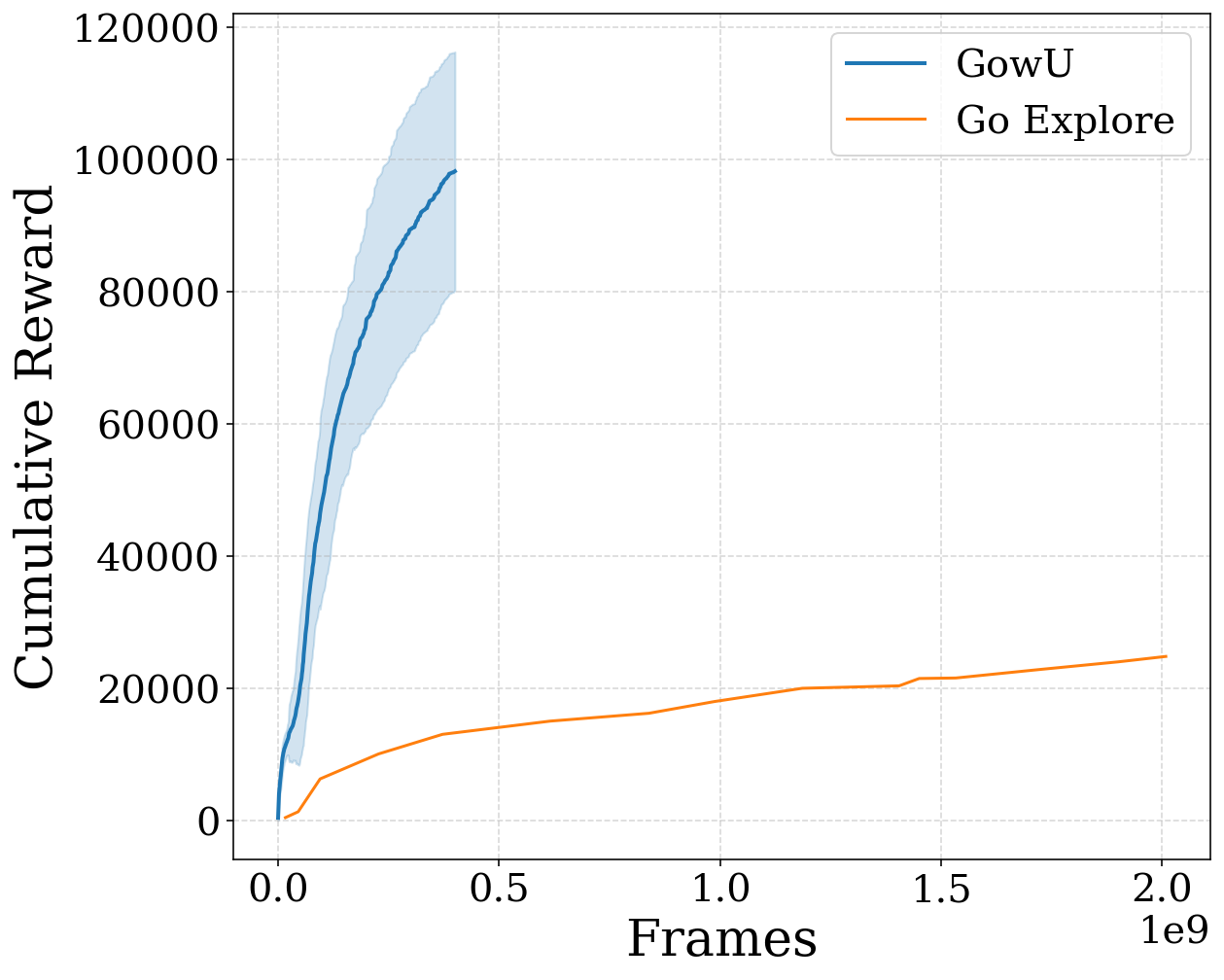} &
        \includegraphics[width=0.31\linewidth, trim=0pt 0pt 0pt 0pt, clip]{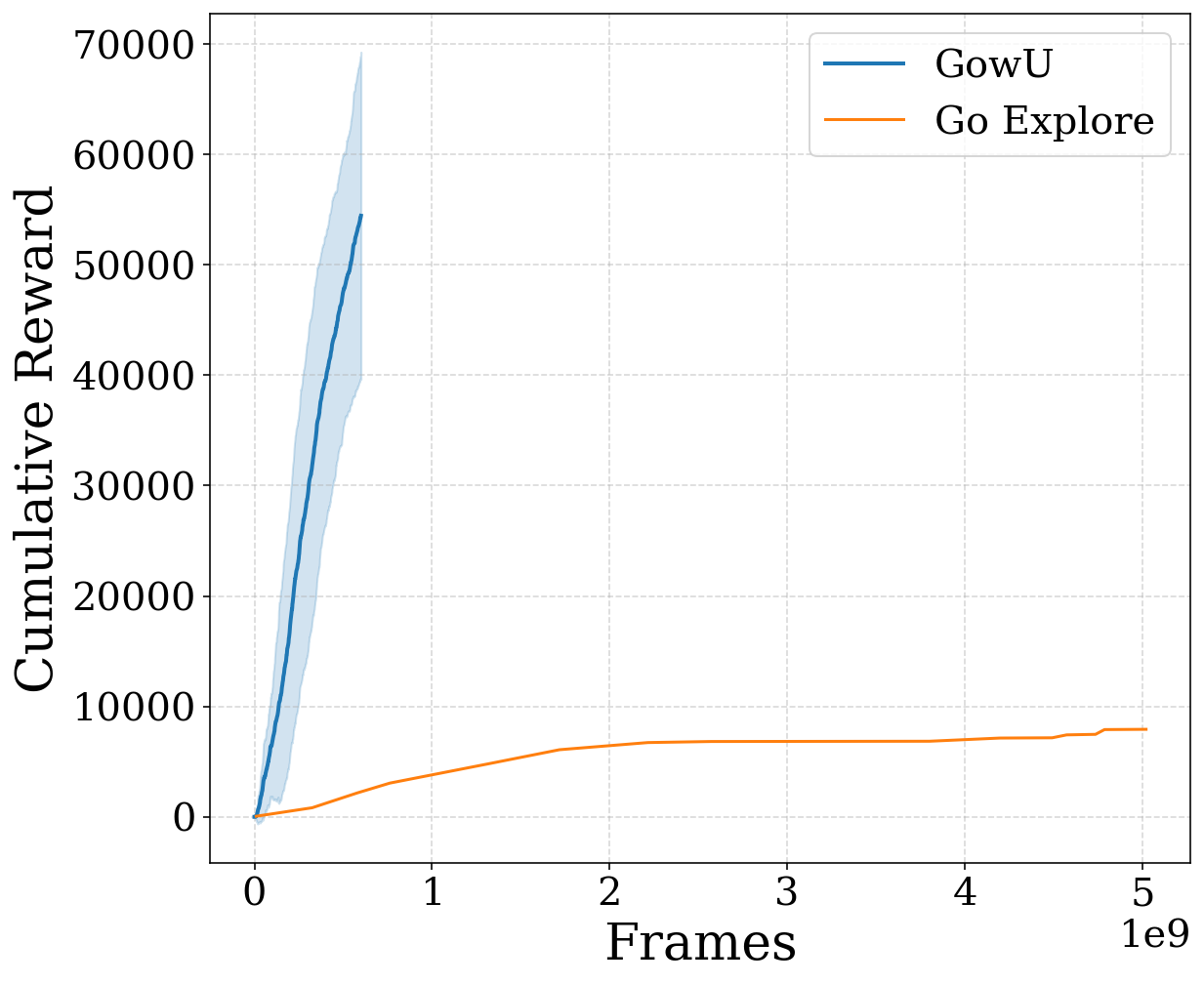} &
        \includegraphics[width=0.31\linewidth, trim=0pt 0pt 0pt 0pt, clip]{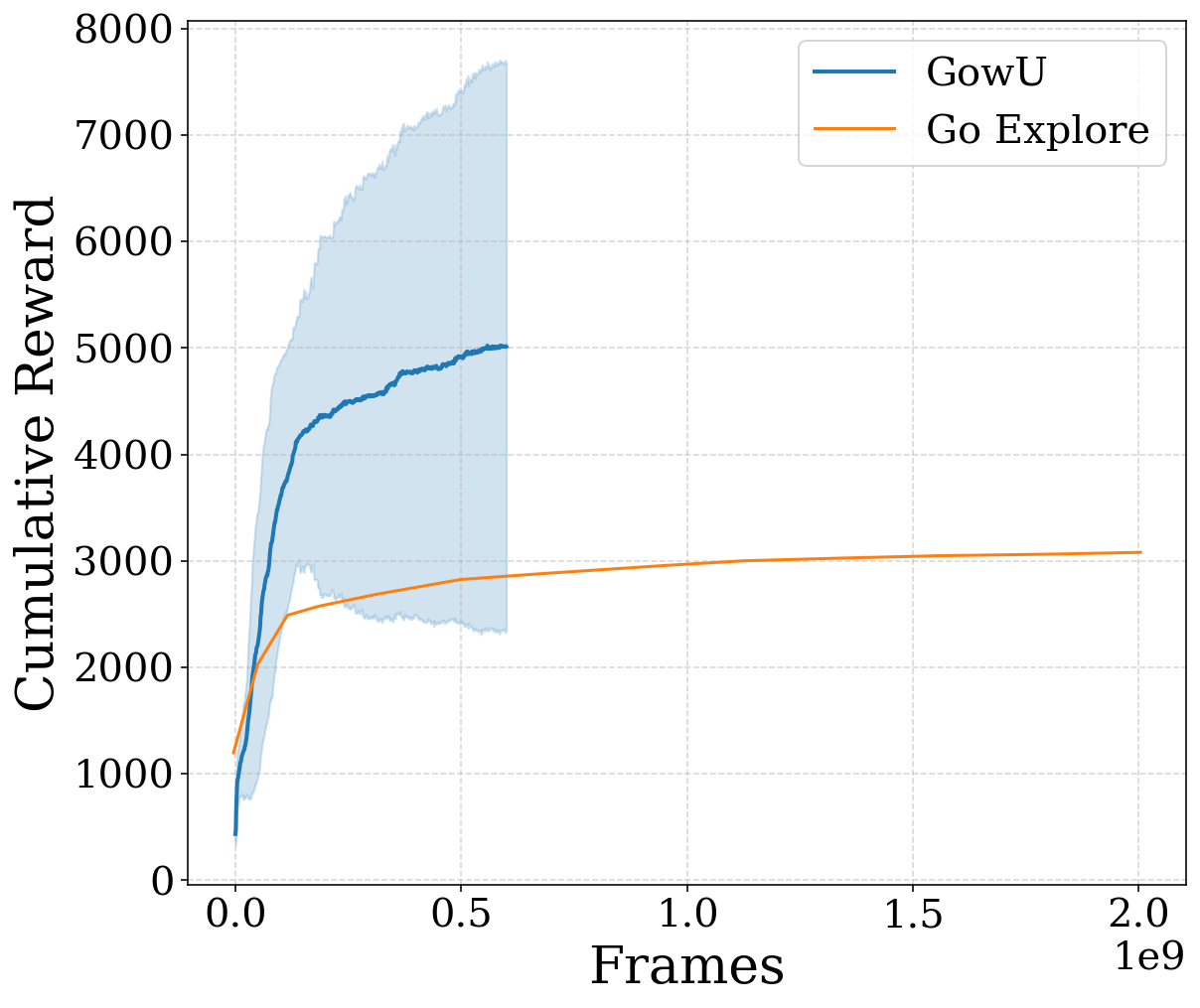} \\
        {\small (a) \montezuma} & {\small (b) \pitfall} & {\small (c) \venture}
    \end{tabular}}
    \caption{Phase~I exploration on Atari: \gowu vs.\ \goexplore \citep{ecoffet2021first}. Mean cumulative reward ($\pm$ std) across 100 seeds as a function of game frames. \goexplore curves are approximated from Extended Data Fig.~2 in \citep{ecoffet2021first}.}
    \label{fig:atari_exploration}
\end{figure}

We run \gowu for 100 seeds on each Atari game---400M frames per seed for \montezuma and 600M for \pitfall and \venture---to evaluate exploration efficiency. \cref{tab:main_comparison} summarizes the mean cumulative reward at the end of exploration alongside the baselines, and \cref{fig:atari_exploration} compares the \gowu and \goexplore \citep{ecoffet2021first} exploration curves on each game. Across all three games, \gowu discovers high-scoring trajectories substantially faster: it reaches higher cumulative rewards within a fraction of the frames required by \goexplore.

We next evaluate whether this efficiency extends to continuous control. \cref{tab:mujoco_exploration} reports statistics on the number of environment frames required for \gowu to discover a task-completing trajectory across $100$ Phase~I seeds for each MuJoCo task. All tasks are solved reliably and efficiently (using substantially fewer frames than in the Atari setting).

\begin{table}[ht]
\centering
\caption{MuJoCo Phase~I exploration efficiency: number of environment frames until a task-completing trajectory is first discovered, across $100$ \gowu seeds.}
\label{tab:mujoco_exploration}
\begin{sc}
\begin{small}
\begin{tabular}{lccccc}
\toprule
\textbf{Task} & \textbf{Mean} & \textbf{Std} & \textbf{Median} & \textbf{Min} & \textbf{Max} \\
\midrule
Hammer   & 26{,}797  & 13{,}983  & 24{,}142  & 4{,}713     & 73{,}306 \\
Door     & 272{,}688 & 200{,}824 & 224{,}338 & 17{,}086    & 943{,}066 \\
Relocate & 275{,}372 & 839{,}815 & 69{,}546  & 2{,}137     & 8{,}015{,}087 \\
AntMaze  & 10{,}161{,}272 & 6{,}697{,}736 & 8{,}517{,}186 & 3{,}078{,}412 & 48{,}332{,}904 \\
\bottomrule
\end{tabular}
\end{small}
\end{sc}
\end{table}

\subsection{Phase II Results: Policy Distillation}
Following the exploration phase, we distill the discovered trajectories into deployable policies via the backward learning curriculum described in \cref{sec:backwardimplement}. For each task, we conduct 10 Phase~II runs, where each run trains on a unique set of 10 demonstrations, each extracted from a different \gowu exploration run. Each Phase~II run is repeated with 5 random seeds, and its result is averaged over the seeds. We note that unlike the demonstrations used for backward learning in \citep{salimans2018learning,ecoffet2021first}, which consist of a checkpoint at every environment step, ours consist of checkpoints spaced \innersteps apart, since we only expand the tree at the end of each rollout (see \cref{alg:population_iteration}). Despite the gaps between checkpoints, the backward algorithm successfully learns high-performing policies.

\begin{table}[H]
    \centering
    \caption{Detailed Phase~II robustification statistics for \gowu on Atari. We conduct 10 Phase~II runs per game, where each run trains on a unique set of 10 demonstrations, each from a different exploration run. Each Phase~II run is repeated with 5 random seeds, and its result is averaged over the seeds. We report the mean, standard deviation, median, worst-run, and best-run policy scores across the 10 runs.}
    \label{tab:robustification_stats}
    \begin{sc}
        \begin{small}
            \begin{tabular}{lccccc}
                \toprule
                \textbf{Game} & \textbf{Mean} & \textbf{Std} & \textbf{Median} & \textbf{Worst-Run} & \textbf{Best-Run} \\
                \midrule
                Montezuma     & 196{,}312     & 49{,}670     & 189{,}151       & 130{,}501          & {\bf 263{,}111}   \\
                Pitfall       & 101{,}972     & 2{,}226      & 102{,}296       & 97{,}090           & {\bf 105{,}589}   \\
                Venture       & 6{,}436       & 1{,}504      & 6{,}177         & 4{,}743            & {\bf 9{,}357}     \\
                \bottomrule
            \end{tabular}
        \end{small}
    \end{sc}
\end{table}

For Atari, the robustification column of \cref{tab:main_comparison} reports the mean policy scores ($\pm$ standard deviation) across the 10 Phase~II runs; full statistics are provided in \cref{tab:robustification_stats}. \gowu's distilled policies surpass all baselines on every game by a wide margin. For \montezuma, the best run achieves $263{,}111$. Across all games, even the worst-run scores remain substantially above the baselines.

For the MuJoCo tasks, we evaluate the distilled policies by their success rate: $100.0\% { \pm 0.1}$ on \texttt{hammer}, $98.1\% { \pm 0.6}$ on \texttt{door}, $95.2\% { \pm 1.7}$ on \texttt{relocate}, and $96.1\% { \pm 1.9}$ on AntMaze (see \cref{tab:mujoco_robustification_stats} for full statistics). To the best of our knowledge, no prior method has achieved comparable success rates on these tasks from pixel observations with sparse rewards, without expert demonstrations (see \cref{app:additional_related} for a detailed discussion of existing approaches).
 
\begin{table}[H]
    \centering
    \caption{Phase~II success rate statistics for MuJoCo tasks. We conduct 10 Phase~II runs per task, where each run trains on a unique set of 10 demonstrations and is repeated with 5 random seeds, and its result is averaged over the seeds. We report the mean, standard deviation, median, worst-run, and best-run success rates across the 10 runs.}
    \label{tab:mujoco_robustification_stats}
    \begin{sc}
        \begin{small}
            \begin{tabular}{lccccc}
                \toprule
                \textbf{Task} & \textbf{Mean} & \textbf{Std} & \textbf{Median} & \textbf{Worst-Run} & \textbf{Best-Run} \\
                \midrule
                Hammer        & 1.000         & 0.001        & 1.000           & 0.997              & {\bf 1.000}       \\
                Door          & 0.981         & 0.006        & 0.981           & 0.972              & {\bf 0.992}       \\
                Relocate      & 0.952         & 0.017        & 0.947           & 0.933              & {\bf 0.973}       \\
                AntMaze       & 0.961         & 0.019        & 0.964           & 0.916              & {\bf 0.980}       \\
                \bottomrule
            \end{tabular}
        \end{small}
    \end{sc}
\end{table}
 We report the Phase~II frame budgets in \cref{app:backward_eval}; in all cases, the cost of policy distillation far exceeds that of Phase~I, making exploration, with our approach, effectively the easier part.

\subsection{Ablation Study}
We conduct ablation experiments on \montezuma to validate the key design choices of \gowu (see \cref{sec:ablations} for details). Four findings stand out: (1)~removing the uncertainty-based winner selection causes exploration to fail, confirming that uncertainty is essential for directing the search; (2)~disabling group consolidation also degrades performance, though the variant still performs better than the no-uncertainty ablation; (3)~we vary the number of parallel groups $M \in \{1, 2, 4, 8\}$ while keeping the total particle count fixed---$2$ and $4$ groups yield similar performance, while a single group is the worst and $8$ groups degrades due to too few particles per group; and (4)~\gowu randomizes its search hyperparameters at each step by default; comparing this against fixing them for the entire run confirms that per-iteration randomization provides robust performance without requiring careful tuning.

\subsection{Discussion and Future Work}
\label{sec:discussion}
 
Our findings suggest that \textsc{GWTW}-style particle-based tree search guided by a measure of uncertainty is a highly promising paradigm for solving hard-exploration tasks. Although particle-based methods that search directly in state space have received limited attention in the context of deep exploration in Atari and robotics simulation, they have gained significant traction in the LLM setting. A growing body of work uses populations of partial generations---steered via resampling---to guide language model outputs toward desired properties \citep{lew2023sequential,zhao2024probabilistic,feng2024step,lipkin2025fast,loula2025syntactic,grand2025self,puri2025probabilistic}. \citet{golowich2025reject} frame these methods as Sequential Monte Carlo and provide theoretical guarantees. In these approaches, a process reward model typically drives the resampling step; an interesting direction would be to replace it with an uncertainty-based signal, as in \gowu. More broadly, the LLM setting is especially well-suited to our framework's reset primitive, since the ``environment'' is a token sequence and resetting merely entails restoring a context window \citep{foster2025goodfoundationnecessaryefficient,chang2024datasetresetpolicyoptimization}.

In another direction, by bypassing the need for shaped rewards or expert demonstrations, \gowu could open the door to scaling up open-ended robotic learning in simulation \citep{openendedlearningteam2021openendedlearningleadsgenerally}, where resets are readily available and leveraging them effectively is crucial \citep{mhammedi2024power}.

On the methodological side, an important direction is the design of more principled uncertainty measures. \rnd is scalable and effective in our setting \citep{burda2018exploration}, but can still assign persistently high novelty to stochastic observations \citep{mavor2022stay}. Promising alternatives include learned representations that explicitly disentangle aleatoric and epistemic uncertainty \citep{jarrett2023curiosityhindsightintrinsicexploration}, as well as temporal contrastive features that capture temporal structure and may provide a stronger basis for noise-robust exploration (see \cref{app:additional_related}).
\clearpage 

\icml{
\section*{Acknowledgments}
We thank Arkanath Pathak for discussions during the early stages of this project.
}
\neurips{
\begin{ack}
We thank Arkanath Pathak for discussions during the early stages of this project.
\end{ack}
}
\arxiv{
\section*{Acknowledgments}
We thank Arkanath Pathak for discussions during the early stages of this project.
}

\bibliography{references}
\icml{\bibliographystyle{icml2026}}
\neurips{\bibliographystyle{plainnat}}
\arxiv{\bibliographystyle{plainnat}}
\clearpage

\appendix
\crefalias{section}{appendix}

\etocdepthtag.toc{appendix}
{\setcounter{tocdepth}{2}
\etocsettagdepth{main}{none}
\etocsettagdepth{appendix}{subsection}
\etocsettocstyle{\subsection*{Appendix Table of Contents}}{}
\tableofcontents
}
\vspace{1em}

\clearpage
\iftoggle{neurips}{\section{Algorithm Pseudocode}
\label{app:algorithms}

This section presents the full pseudocode for \gowu and introduces the notation used therein.

\paragraph{Notation} We define the lexicographic order $\succ^{\lex}$ over tuples such that $(b, c) \succ^{\lex} (b', c')$ if and only if $b > b'$ or ($b = b'$ and $c > c'$). Accordingly, for any two functions $f$ and $g$, we use the notation $\operatorname*{lex-argmax}_{u} (f(u), g(u))$ to denote the argument $u$ that maximizes the vector objective with respect to $\succ^{\lex}$. This operator prioritizes maximizing the first component $f(u)$, using the second component $g(u)$ strictly as a tie-breaker. The pseudocode also uses a reset primitive, $\text{Reset}(p_{\text{target}}, p_{\text{source}})$, which sets the internal state of $p_{\text{target}}$ to match $p_{\text{source}}$, allowing $p_{\text{target}}$ to resume trajectory generation from the state of $p_{\text{source}}$.

\begin{algorithm}[ht]
\caption{\evolvegroup}
\label{alg:population_iteration}
\begin{algorithmic}[1]
\item[] \textbf{input:} Population $P=\{p_1, \dots, p_N\}$, lineage tree $\mathcal{T}$, uncertainty $U$
\item[] \textbf{params:} \outersteps, \innersteps, and \rollback ranges: $[K_{\min},K_{\max}]$, $[T_{\min},T_{\max}]$, $[k_{\min},k_{\max}]$
\vspace{0.2cm}
\STATE Sample \outersteps $K \sim \text{Unif}(K_{\min}, K_{\max})$
\FOR{$k=1, \dots, K$}
    \FOR{each particle $p_i \in P$}  
        \STATE Sample \innersteps $T_i \sim \text{Unif}(T_{\min}, T_{\max})$   \mline{line:sample}
        \STATE  \algcommentbiglight{Rollout particle and update uncertainty}
        \FOR{$t = 1, \dots, T_i$} \label{line:step}
            \STATE $U \gets p_i.\step(1, U)$ \hfill \algcommentlight{Execute one step}
            \STATE \textbf{if} {$p_i$ is \dead or $p_i.R$ increased} \textbf{then} \textbf{break} \hfill\algcommentlight{Halting on reward does not kill the particle} \mline{line:rewardbreak}
        \ENDFOR
        \IF{$p_i$ is \textsc{alive}}
            \STATE $\mathcal{T}.\addchild(p_i)$\hfill \algcommentlight{Expand tree for surviving particles} \mline{line:addchild}
            \ENDIF
    \ENDFOR

    \IF{All $p_i \in P$ are \textsc{Dead}} \mline{line:alldead}
        \STATE  \algcommentbiglight{Failure recovery via rollback}
        \STATE $v_{i,\anc} \gets p_i.\getancestor(k_{\min}, k_{\max})$, $\forall i$ \hfill \algcommentlight{Pick $k_i$th ancestor of $p_i$ with $k_i\sim \text{Unif}(k_{\min},k_{\max})$, $\forall i$} 
        \STATE $P \gets \{ \reset(p_i, v_{i,\anc}) \mid \forall i \}$ \hfill \algcommentlight{Restore all particles to ancestor checkpoints} \mline{line:rollback}
    \ELSE
        \STATE  \algcommentbiglight{Uncertainty-aware redistribution}
        \STATE $p_{\winner} \gets \lexargmax_{p \in P, p.\textsc{alive}} (p.R, U(p))$ \hfill \algcommentlight{Best alive: max reward, break ties by uncertainty} \mline{line:winner}
        \FOR{each $p_i \in P$}
            \IF{$p_i$ is \textsc{Dead}}
                \STATE $\reset(p_i, p_{\winner})$ \hfill \algcommentlight{Prune \& respawn; $p_i$ becomes \alive after the reset} \mline{line:prune}
            \ENDIF
        \ENDFOR
    \ENDIF
\ENDFOR
\STATE  \algcommentbiglight{Group consolidation (lexicographic ranking)}
\STATE $p_{\winner} \gets \lexargmax_{p \in P} (p.R, U(p))$ \mline{line:groupwinner}
\STATE  \mline{line:single}$P \gets \{ \reset(p_j, p_{\winner}) \mid \forall j \}$ \hfill \algcommentlight{Reset all particles to winner's checkpoint}
\STATE \textbf{return} $P,\mathcal{T}, U, p_{\winner}$
\end{algorithmic}
\end{algorithm}

\begin{algorithm}[ht]
\caption{\gowu: \textsc{Go-With-Uncertainty}}
\label{alg:parallel_evolution}
\begin{algorithmic}[1]  
\item[] \textbf{input:} Env.~$\mathcal{E}$, iterations $N_{\text{iter}}$, number of groups $M$, and number of particles per group $N$
\vspace{0.2cm}
\STATE Initialize lineage tree $\mathcal{T}$ with root node $s_0$ (initial state)
\STATE Initialize groups $\{G_1, \dots, G_M\}$, each with $N$ particles, all starting at $s_0$
\STATE Initialize shared uncertainty estimator $U$
\STATE Initialize $p_{\winner}$ at state $s_0$ with cumulative reward $0$
\FOR{$i=1, \dots, N_{\text{iter}}$}
\FOR{each Group $G_m$ in parallel}
\STATE \algcommentbiglight{Global synchronization}
\STATE $p_{m} \gets \lexargmax_{p \in G_m} (p.R, U(p))$ \hfill \IF{$p_{m}.R < p_{\winner}.R$}
             \STATE Reset $G_m$ to the state of $p_{\winner}$ \hfill \algcommentlight{Restore all particles to global winner's checkpoint} \mline{line:resetgroup}
        \ENDIF
    \STATE \algcommentbiglight{Evolve group}
        \STATE $G_m, \mathcal{T}, U, p_{m} \gets\evolvegroup(G_m, \mathcal{T}, U)$\hfill \algcommentlight{\cref{alg:population_iteration}}
    \ENDFOR
\STATE \algcommentbiglight{Update global winner}
\STATE $p_{\winner} \gets \lexargmax_{p \in \{p_1, \dots, p_M, p_{\winner}\}} (p.R, U(p))$ \mline{line:identify}
\ENDFOR
\end{algorithmic}
\end{algorithm}

    \begin{figure}[ht]
    \centering
        \includegraphics[width=\linewidth, trim=0pt 90pt 0pt 50pt, clip]{figures/gowu_schematic.pdf}
        \caption{Schematic overview of \gowu with a single group of particles. The algorithm maintains a population of particles ($p_1, p_2, p_3$) that explore the state space via multi-step rollouts. During an outer step, if a particle reaches a \dead state (e.g., $p_2$ on level one), it is pruned and its state is reset via a \emph{Reset} to the winner---the \alive node maximizing accumulated reward, with epistemic uncertainty used as a tie-breaker. After $K$ outer steps, a \emph{Group Consolidation Reset} syncs all particles, including \alive particles, to the current winner.}
        \label{fig:gowu_schematic}
        \end{figure}

}{}

\section{Implementation Details for \gowu}
\label{app:implementation}
\label{app:distributed}

This appendix provides full implementation details for the distributed architecture summarized in \cref{sec:implementation}, including the uncertainty estimator architecture and training procedure. The system decouples environment simulation, population management, and uncertainty model training, enabling high-throughput exploration. It consists of three logical components communicating asynchronously: a \emph{central coordinator}, a pool of \emph{distributed rollout workers}, and a dedicated \emph{\rnd learner}.

\subsection{Central Coordinator}

The central coordinator acts as the orchestrator of the search. It manages the global state of the population and coordinates the branches of exploration. Its responsibilities are as follows:

\paragraph{Population state management}
The coordinator tracks the status of all $M \times N$ particles, including each particle's current environment checkpoint, cumulative reward, and survival status (\alive or \dead, as defined in \cref{sec:preliminaries}).

\paragraph{State-lineage tree maintenance}
The coordinator hosts the global state-lineage tree (\cref{sec:lineage_tree}) and is responsible for expanding it with new nodes as workers report completed rollouts. Each node stores the environment checkpoint, cumulative reward, and a parent pointer. To extract a demonstration path for Phase~II, we select the highest-reward leaf and traverse the parent pointers upward, collecting the environment checkpoint and cumulative reward at each node.

\paragraph{The Go-With-The-Winner loop}
The coordinator drives the top-level iterations of the \gwtw loop (\cref{alg:parallel_evolution}). Each iteration proceeds in three phases:
\begin{enumerate}[nosep, leftmargin=*]
\item \textbf{Dispatch:} The coordinator issues remote procedure calls (RPCs) to trigger rollouts on the worker pool, providing each worker with the starting checkpoint and group index.
\item \textbf{Aggregate:} It waits for all parallel rollouts to complete and collects the resulting endpoints, trajectories, and rewards.
\item \textbf{Sync and prune:} The coordinator applies the winner-selection and pruning logic (\cref{sec:reassignment}) and instructs the workers to overwrite failing particles with clones of the winners.
\end{enumerate}

\subsection{Distributed Rollout Workers}

The workers are parallel execution nodes that drive environment interactions. They act as stateless clients that execute directives issued by the coordinator.

\paragraph{State restoration}
Each worker receives a compact environment snapshot (checkpoint) from the global lineage tree and restores the local simulator to that exact state before starting a rollout.

\paragraph{Rollout execution}
The worker steps through the environment for the prescribed number of steps. Each particle commits to a single random action, held fixed for the duration of the rollout segment (see \cref{sec:implementation}). In the discrete setting, this action is played with probability $1 - \varepsilon$ ($\varepsilon = 0.2$), with a uniform random action played otherwise. In the continuous setting, the committed action is played deterministically throughout the segment.

\paragraph{Back-reporting}
Upon completing the rollout, the worker transmits the endpoint environment checkpoint and cumulative reward back to the coordinator. The coordinator uses these to create a new node in the state-lineage tree (\cref{sec:lineage_tree}), which stores only the endpoint checkpoint, cumulative reward, and a parent pointer.

\subsection{RND Learner}
\label{app:uncertaintyrnd}

The uncertainty oracle $U$ is instantiated using standard \rnd \citep{burda2018exploration}, which quantifies novelty via the prediction error of a trained network against a fixed, random target. We choose \rnd for its scalability and efficiency compared to ensemble-based uncertainty methods \citep{osband2016deep,lakshminarayanan2017simple}. The module consists of a fixed \textit{target network} and a trainable \textit{predictor network}. Both use standard Atari convolutional torsos \citep{mnih2015human} followed by MLP heads consisting of two 1024-unit linear layers separated by a ReLU activation. We use orthogonal initialization for the torso weights and normalize inputs (running mean/std) clipped to $[-5, 5]$, consistent with standard \rnd implementations \citep{burda2018exploration}. The predictor is trained to minimize the Mean Squared Error (MSE) against the target embedding.

\paragraph{Replay buffer}
Workers stream their observations into a shared replay buffer of capacity $512{,}000$. Each entry has the form $(s, a)$, where $s$ is the observation and $a$ the action.

\paragraph{Asynchronous training}
The \rnd learner continuously samples mini-batches from the shared replay buffer and runs gradient-descent steps to update the predictor network, using a batch size of 128 and Adam with a learning rate of $3 \times 10^{-4}$. The coordinator periodically fetches the latest predictor weights from the learner to score particle states.

\subsection{Population Management Hyperparameters}
\label{sec:hyperparams}
\cref{tab:hyperparams} summarizes the main population management hyperparameters introduced in \cref{sec:method}. 

\begin{table}[h] 
\centering
\caption{Population management hyperparameters for \gowu. The \textbf{MuJoCo} column applies to all continuous-control tasks (AntMaze and Adroit).}
\label{tab:hyperparams}
\begin{sc}
\begin{small}
\begin{tabular}{lcccc}
\toprule
\textbf{Parameter} & \textbf{Montezuma} & \textbf{Pitfall} & \textbf{Venture} & \textbf{MuJoCo} \\
\midrule
Groups ($M$)             & 4     & 4     & 4      & 16    \\
Particles per group ($N$) & 32   & 32    & 32     & 8    \\
\outersteps range ($[K_{\min}, K_{\max}]$) & $[10, 20]$ & $[10, 20]$ & $[6, 16]$  & $[20, 40]$ \\
\innersteps range ($[T_{\min}, T_{\max}]$)      & $[5, 15]$ & $[10, 20]$ & $[10, 20]$ & $[3, 4]$  \\
\rollback range $[k_{\min}, k_{\max}]$       & [1, 3]    & [1, 4]     & [1, 12]      & [1,20]     \\
\bottomrule
\end{tabular}
\end{small}
\end{sc}
\end{table}

\subsection{Computational Resources}
\label{app:compute}

\Cref{tab:compute} reports wall-clock runtime, exploration score, and the number of \rnd SGD steps for Phase~I of \gowu on \montezuma, averaged over 100 seeds. Each run uses the distributed architecture described above with 1 TPU for the \rnd uncertainty estimator, 128 CPUs (one per rollout worker), and 1 CPU for the coordinator.

\begin{table}[ht]
    \centering
    \caption{Computational cost of \gowu Phase~I on \montezuma, averaged over 100 seeds.}
    \label{tab:compute}
    \resizebox{\columnwidth}{!}{\begin{tabular}{lcccc}
            \toprule
            \textbf{Metric}       & \textbf{100M frames}            & \textbf{200M frames}            & \textbf{300M frames}            & \textbf{400M frames}              \\
            \midrule
            Clock runtime (hours) & $4.53 \pm 1.84$                & $6.98 \pm 1.92$                & $9.01 \pm 2.58$               & $10.71 \pm 3.35$                 \\
            Mean score             & $48{,}444 \pm 14{,}417$         & $76{,}014 \pm 16{,}626$         & $89{,}485 \pm 18{,}760$         & $98{,}249 \pm 18{,}102$           \\
            \rnd SGD steps        & $0.77\text{M} \pm 0.28\text{M}$ & $1.38\text{M} \pm 0.39\text{M}$ & $1.86\text{M} \pm 0.57\text{M}$ & $2.28\text{M} \pm 0.81\text{M}$ \\
            \bottomrule
        \end{tabular}}
\end{table}

Notably, the total number of \rnd SGD steps at 400M frames (${\sim}2.3\text{M}$) is comparable to the ${\sim}3\text{M}$ model update steps used by intrinsic-motivation baselines such as \textsc{BYOL-Explore} \citep{guo2022byol} and \textsc{BYOL-Hindsight} \citep{jarrett2023curiosityhindsightintrinsicexploration}, indicating no additional computational overhead from the uncertainty estimator relative to standard deep RL training.

\section{Implementation Details for the Backward Algorithm}
\label{app:backward}
Here, we describe our implementation of the backward algorithm. We first explain how demonstrations are decomposed into segments and then detail the curriculum learning strategy applied to each segment. Finally, we provide the specific configuration parameters and hyperparameters used in our experiments.

\subsection{Overview and segmentation}
Our approach builds on the backward learning algorithm of \citet{salimans2018learning}, as also used in the robustification phase of \goexplore \citep{ecoffet2021first}, and extends it with several modifications. The agent learns to solve the task by starting from states near the end of a demonstration and gradually moving the starting point backwards. To break a single long backward trajectory into smaller, more manageable chunks---making training faster and more stable---we decompose each demonstration trajectory $\tau = \{s_0, a_0, r_0, \dots, s_T\}$ into at most $K_{\max}$ segments delineated by reward events. Concretely, we first identify all timesteps at which the agent receives a non-zero reward. If there are more such timesteps than $K_{\max}$, we evenly downsample them to obtain exactly $K_{\max}$ boundary points. Let $T_k$ denote the $k$-th selected boundary; segment~$k$ then covers the portion of the trajectory culminating at $T_k$. For Atari, we set $K_{\max} = 10$; for the MuJoCo tasks (Adroit and AntMaze), we set $K_{\max} = 1$, treating the entire demonstration as a single segment where the starting state is progressively moved back toward the initial state. Our implementation supports training on multiple demonstrations simultaneously: each demonstration is segmented independently. Importantly, the resulting segments are not simply aggregated into a flat pool; the system tracks the ordering of segments and which demonstration each segment originates from, since each segment's curriculum window is allowed to extend backward beyond the segment's own start boundary.

\paragraph{Distributed architecture} The system uses a distributed actor-learner architecture. A centralized \emph{curriculum server} maintains the global curriculum state---tracking the current start index, and success rates for every segment---and serves starting states to the actors. Multiple distributed \emph{actors} concurrently fetch start states from the curriculum server, execute episodes, and stream trajectories to a centralized \emph{learner}, which continuously updates the PPO policy weights. Background \emph{evaluators} run asynchronously, periodically fetching the latest weights to assess performance (see \cref{app:backward_eval}). This architecture allows the algorithm to learn different stages of the task in parallel, scaling efficiently with the number of actors.

\subsection{Curriculum strategy}
For a chosen segment ending at $T_{\text{end}}$, the curriculum maintains a \textit{current start index} $t_{\text{curr}}$ (initially set to $T_{\text{end}}$) and a fixed window size $\Delta$. This index represents the latest point in the trajectory from which the agent is currently learning to reach the segment's goal. The curriculum dynamically adjusts the start position based on the agent's performance.

\paragraph{State sampling}
At the beginning of an episode, a start index $t_{\text{start}}$ is sampled uniformly from the interval $[t_{\text{curr}} - \Delta, t_{\text{curr}}]$. The environment is reset to the state $s_{t_{\text{start}}}$ from the demonstration.

\paragraph{Success criteria}
An episode is considered successful if the agent achieves a return $R_{\text{agent}}$ comparable to the return of the demonstration segment from $t_{\text{start}}$ to $T_{\text{end}}$, denoted as $R_{\text{demo}}(t_{\text{start}})$. Success is strictly defined as:
\[
R_{\text{agent}} \ge R_{\text{demo}}(t_{\text{start}}) - \epsilon_{\text{tol}}
\]
where $\epsilon_{\text{tol}}$ is a tolerance parameter (see \cref{tab:backward_config} for environment-specific values). An episode is terminated early if the number of steps exceeds $\mu \cdot \bar{L} + b$, where $\bar{L}$ is an exponential moving average (with smoothing factor $\alpha = 0.9$) of successful rollout lengths, $\mu = 2.0$ is a multiplier, and $b$ is a fixed buffer ($b = 500$ by default; $b = 100$ for Adroit). For Atari, training episodes are additionally terminated early upon loss of life; for AntMaze, episodes are terminated if the ant flips over (see \cref{app:mujoco_rewards}).

\paragraph{Curriculum progression}
The curriculum tracks the success rate $S$ over a buffer of the most recent $N_{\text{update}}$ rollouts. Two distinct success thresholds govern progression:
\begin{itemize}[nosep, leftmargin=*]
    \item \textbf{Regression (moving backwards):} If $S \ge S_{\text{req}}$ (default $0.2$), the window moves backward to include earlier states. The current start index is updated as $t_{\text{curr}} \leftarrow \max(0, t_{\text{curr}} - \delta_{\text{back}})$, where the step size $\delta_{\text{back}}$ is sampled uniformly from $[\alpha_{\text{dec}} \cdot \Delta,\; \beta_{\text{dec}} \cdot \Delta]$. Here, $\alpha_{\text{dec}}$ and $\beta_{\text{dec}}$ are decrease multipliers that control the step size range (see \cref{tab:backward_config}). This stochastic step size prevents the curriculum from getting stuck in local cycles. Note that the lower bound is the absolute beginning of the demonstration (the initial state), not the beginning of the segment; this means the curriculum window for a given segment can extend well beyond the segment's own start boundary, and each segment's curriculum eventually requires the agent to achieve the segment's reward target starting from the very beginning of the demonstration.
    \item \textbf{Simplification (moving forwards):} If $S < S_{\text{req}}$, the task is deemed too difficult, and the window moves forward. The update is $t_{\text{curr}} \leftarrow \min(T_{\text{end}}, t_{\text{curr}} + \delta_{\text{fwd}})$, where the step size $\delta_{\text{fwd}}$ is sampled uniformly from $[\alpha_{\text{inc}} \cdot \Delta,\; \beta_{\text{inc}} \cdot \Delta]$, with $\alpha_{\text{inc}}$ and $\beta_{\text{inc}}$ being the increase multipliers.
    \item \textbf{Completion criterion:} When $t_{\text{curr}}$ reaches step~$0$ (the beginning of the demonstration), a stricter threshold $S_{\text{req,begin}} = 0.95$ must be met before the segment is considered solved. This ensures the agent can reliably execute the full trajectory from the very start of the demonstration to the segment's reward boundary.
\end{itemize}

\subsection{Environment-specific configurations}
\label{app:backward_envconfig}

The backward curriculum is configured differently across environment families:

\paragraph{Atari (Montezuma's Revenge, Pitfall!, Venture)}
Demonstrations are decomposed into up to $K_{\max} = 10$ segments. The window size is $\Delta = 25$, and the decrease/increase multipliers are $(\alpha_{\text{dec}}, \beta_{\text{dec}}) = (0.25, 0.75)$ and $(\alpha_{\text{inc}}, \beta_{\text{inc}}) = (0.5, 1.0)$, respectively. For \montezuma and \venture, reward clipping is applied (bounding cumulative targets to $[-1, 1]$) with a strict tolerance $\epsilon_{\text{tol}} = 0$. For \pitfall, rewards are scaled by $0.001$ (without clipping) and the tolerance is relaxed to $\epsilon_{\text{tol}} = 1500$ to accommodate the game's scoring structure, where the agent can lose points. For \montezuma, the agent's observations include the frame at which the last reward was received, providing context when the same room appears multiple times in a demonstration. Additionally, after successfully achieving the segment's reward target, the agent is allowed to continue playing for a random number of extra frames (up to $\sim e^7 \approx 1096$); this provides additional exploration beyond the target state.

\paragraph{Adroit (door, hammer, relocate)}
The entire demonstration is treated as a single segment ($K_{\max} = 1$). The window size is reduced to $\Delta = 3$, and all multipliers are set deterministically to $1/3$, removing the stochastic step-size variation. Intermediate contact rewards from Phase~I are discarded; only task-completion events are used as reward signals. During the backward curriculum, the goal configuration (i.e., the randomized target location and object placement) is held fixed to the one used during exploration. Whenever an actor samples a start index of $0$ from the curriculum window, the episode uses a clean environment reset (i.e., a true \texttt{env.reset()}, which randomizes the goal configuration). This allows the policy to begin generalizing across target locations and object placements. 

\paragraph{AntMaze}
As with Adroit, the demonstration is treated as a single segment ($K_{\max} = 1$). The window size is $\Delta = 25$. An action repeat of $4$ is applied. Clean environment resets are used whenever an actor samples a start index of $0$, as described above for Adroit. The PPO batch size is reduced from $128$ to $64$ (this is mainly to avoid out-of-memory errors).

\subsection{Agent architecture}
We use Proximal Policy Optimization (PPO) as the underlying reinforcement learning algorithm. The policy and value functions use separate encoders based on the IMPALA ResNet architecture \citep{espeholt2018impala}, each followed by fully connected layers. For Atari, each encoder feeds into a dense layer of size $1024$, followed by policy and value heads of size $512$. For Adroit and AntMaze, the architecture is scaled down (to avoid out-of-memory errors): the dense layer is $512$, and the policy and value heads are $256$ each. The encoder input varies by environment; see \cref{app:obs_processing} for details on the observation processing pipeline for each environment family.

\subsection{Hyperparameters and configuration}

For each demonstration, we run on 5 random seeds.

\begin{table}[h]
    \centering
    \caption{PPO Hyperparameters (defaults; see \cref{app:backward_envconfig} for environment-specific overrides).}
    \begin{tabular}{l|l}
        \toprule
        Parameter & Value \\
        \midrule
        Optimizer & Adam \\
        Learning Rate & $1 \times 10^{-4}$ \\
        Discount Factor ($\gamma$) & 0.999 \\
        GAE Lambda ($\lambda$) & 0.95 \\
        Unroll Length & 128 \\
        Batch Size & 128 (64 for AntMaze) \\
        Num Epochs & 1 \\
        Num Minibatches & 8 \\
        Entropy Cost & $1 \times 10^{-3}$ \\
        Value Cost & 0.5 \\
        PPO Clipping ($\epsilon$) & 0.1 \\
        Max Gradient Norm & 0.5 \\
        \midrule
        Shared Dense Layer & 1024 (512 for Adroit/AntMaze) \\
        Policy/Value Head & 512 (256 for Adroit/AntMaze) \\
        \bottomrule
    \end{tabular}
\end{table}

\begin{table}[h]
    \centering
    \caption{Backward Algorithm Configuration (defaults; see \cref{app:backward_envconfig} for environment-specific overrides).}
    \label{tab:backward_config}
    \begin{tabular}{l|l}
        \toprule
        Parameter & Value \\
        \midrule
        Required Success Rate ($S_{\text{req}}$) & 0.2 \\
        Completion Threshold ($S_{\text{req,begin}}$) & 0.95 \\
        Start delta window ($\Delta$) & 25 (3 for Adroit) \\
        Min Rollouts for Update ($N_{\text{update}}$) & 32 \\
        Decrease Multipliers ($\alpha_{\text{dec}}, \beta_{\text{dec}}$) & 0.25, 0.75 ($\tfrac{1}{3}$, $\tfrac{1}{3}$ for Adroit) \\
        Increase Multipliers ($\alpha_{\text{inc}}, \beta_{\text{inc}}$) & 0.5, 1.0 ($\tfrac{1}{3}$, $\tfrac{1}{3}$ for Adroit) \\
        Failure Reward Tolerance ($\epsilon_{\text{tol}}$) &  0 (1500 for Pitfall) \\
        EMA Smoothing Factor ($\alpha$) & 0.9 \\
        Early Termination Multiplier ($\mu$) & 2.0 \\
        Rollout Length Buffer ($b$) & 500 (100 for Adroit) \\
        \bottomrule
    \end{tabular}
\end{table}
 
Consistent with \goexplore, we use reward clipping on \montezuma and \venture, and reward scaling on \pitfall.

\subsection{Evaluation}
\label{app:backward_eval}

A background evaluator runs concurrently with training: it periodically fetches the latest policy weights from the learner and executes evaluation rollouts using the deterministic (mode) version of the policy. During training, we run $10$ rollouts per evaluation for Atari and $30$ for MuJoCo, and track the best-performing checkpoint based on these periodic evaluations.
 
\paragraph{Training budget}
For Atari, we train for $15$--$20$B environment frames. For MuJoCo, training continues until all segments have been solved with the $95\%$ completion threshold ($S_{\text{req,begin}}$). \Cref{tab:mujoco_phaseii_budget} reports the total number of environment frames consumed during Phase~II for each MuJoCo task, across 10 runs.

\begin{table}[ht]
    \centering
    \caption{MuJoCo Phase~II training budget: total environment frames consumed per task, across 10 runs (each averaged over 5 seeds).}
    \label{tab:mujoco_phaseii_budget}
    \begin{sc}
        \begin{small}
            \begin{tabular}{lccc}
                \toprule
                \textbf{Task} & \textbf{Median} & \textbf{Std}    & \textbf{Max}    \\
                \midrule
                Hammer        & 9{,}876{,}809   & 2{,}340{,}952   & 19{,}878{,}384  \\
                Door          & 27{,}629{,}359  & 14{,}928{,}478  & 54{,}589{,}532  \\
                Relocate      & 221{,}653{,}220 & 146{,}455{,}143 & 827{,}483{,}942 \\
                AntMaze       & 119{,}160{,}557 & 84{,}845{,}527  & 667{,}819{,}039 \\
                \bottomrule
            \end{tabular}
        \end{small}
    \end{sc}
\end{table}

\paragraph{Final evaluation}
After training is complete, we perform a dedicated evaluation of the best checkpoint identified during training. This checkpoint is evaluated on $500$ rollouts to produce the final policy score.

\paragraph{Evaluation heuristics}
 A well-known exploit in \montezuma allows the agent to remain in a treasure room and collect $1{,}000$-point rewards indefinitely by repeating a specific action sequence, without progressing to the next level. To prevent this from inflating evaluation scores, the evaluator monitors for consecutive rewards of exactly $1{,}000$ points. If $30$ or more such consecutive rewards are detected, the rollout is immediately discarded.

\section{Environment Checkpointing}
\label{app:checkpointing}

Both Phase~I and Phase~II rely on environment resets to restore particles or training episodes to previously visited states. Reconstructing the environment mid-trajectory requires checkpointing both the simulator's physical state and the agent's observation history (e.g., frame stacks). The mechanism differs across environment families:
\begin{itemize}[nosep, leftmargin=*]
    \item \textbf{Atari:} We extract and restore the internal ALE emulator state along with the frame stacker's observation history, following the implementation of \citet[Page~21]{yin2023sampleefficientdeepreinforcement}.
    \item \textbf{Adroit:} We checkpoint the MuJoCo physics state (joint positions and velocities), the random seed (which determines the randomized object and target positions), and task-specific state variables (e.g., whether contact with the target surface has been established).
    \item \textbf{AntMaze:} In addition to the MuJoCo physics state, we save the random seed (which determines the randomized goal coordinates) and the pixel observation stack from the third-person camera renders.
\end{itemize}
In all cases, the frame stack is replaced with the one saved at checkpoint time.

\section{Pitfall Timer}
\label{sec:resets}
Unlike \montezuma and \venture, \pitfall includes a twenty-minute termination timer. We disable this timer during Phase~I to allow non-episodic exploration with \gowu, but final policies are evaluated with the standard timer active. Concretely, we reset the timer at each step by directly writing to the Atari emulator's RAM via \texttt{ale.setRAM()}. The timer is encoded across two RAM bytes: byte~88 stores the minutes and byte~89 stores the seconds. Because the game uses a non-standard internal encoding, we maintain hardcoded lookup tables that map standard minute and second values to the corresponding RAM integers expected by the emulator. We avoid setting the timer to exactly 20:00 or 0:00, as both boundary values trigger unintended side effects in the game state; in practice, we reset it to 19:59. This timer reset is applied only during Phase~I exploration and is disabled during Phase~II and evaluation.

\section{Observation Processing}
\label{app:obs_processing}

This section describes the observation processing pipeline for each environment family. All pipelines produce grayscale, frame-stacked tensors normalized to $[0, 1]$; the specifics differ per domain and are detailed below.

\iftoggle{neurips}{\begin{figure}[ht]
        \centering
        \setlength{\tabcolsep}{4pt}
        \resizebox{\columnwidth}{!}{
            \begin{tabular}{@{}ccc@{}}
                \includegraphics[height=3.4cm]{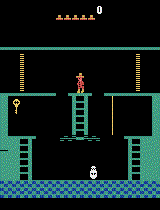} &
                \includegraphics[height=3.4cm]{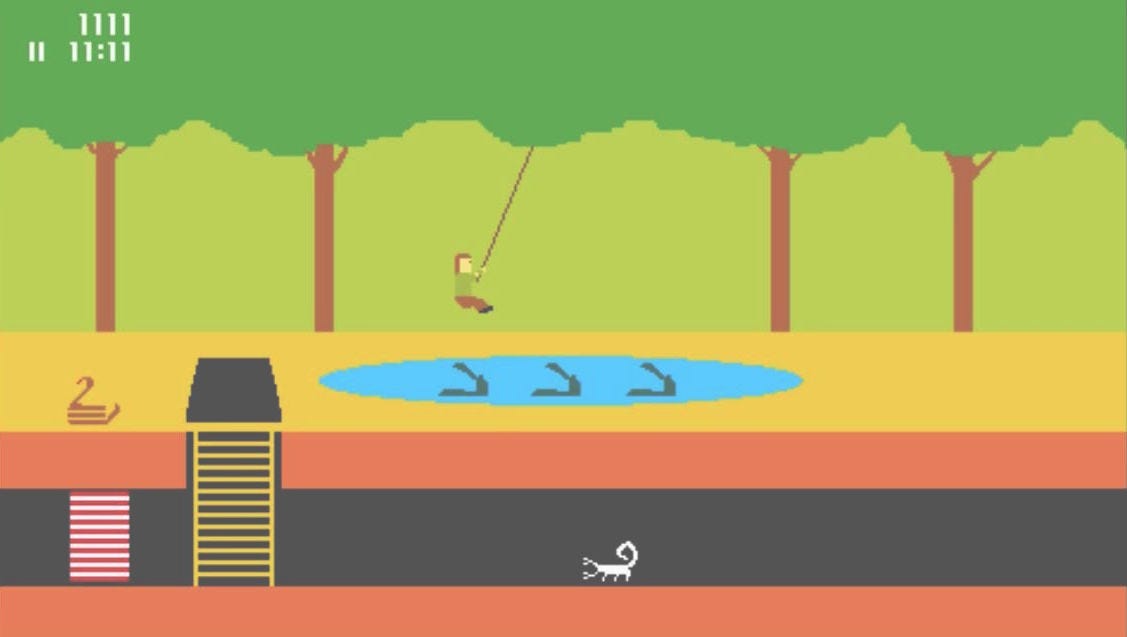}          &
                \includegraphics[height=3.4cm]{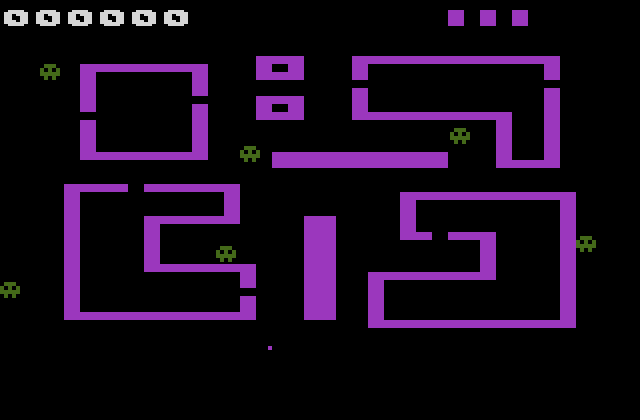}                                                           \\[4pt]
                {\small (a) \montezuma}                                       & {\small (b) \pitfall} & {\small (c) \venture}
            \end{tabular}
        }
        \caption{Fully rendered observations from the three hard-exploration Atari games used in our evaluation.}
        \label{fig:atari_games}
    \end{figure}

    \begin{figure}[t]
        \centering
        \begin{minipage}[b]{0.24\linewidth}
            \centering
            \includegraphics[width=\linewidth, trim=0pt 0pt 0pt 18pt, clip]{figures/door-task.png}\\
            {\small (a) \texttt{door-v0}}
        \end{minipage}
        \hfill
        \begin{minipage}[b]{0.24\linewidth}
            \centering
            \includegraphics[width=\linewidth, trim=0pt 0pt 0pt 18pt, clip]{figures/hammer-task.png}\\
            {\small (b) \texttt{hammer-v0}}
        \end{minipage}
        \hfill
        \begin{minipage}[b]{0.24\linewidth}
            \centering
            \includegraphics[width=\linewidth, trim=0pt 0pt 0pt 18pt, clip]{figures/relocate-task.png}\\
            {\small (c) \texttt{relocate-v0}}
        \end{minipage}
        \hfill
        \begin{minipage}[b]{0.24\linewidth}
            \centering
            \includegraphics[width=\linewidth, trim=0pt 0pt 0pt 18pt, clip]{figures/antmaze-top.png}\\
            {\small (d) AntMaze}
        \end{minipage}
        \caption{MuJoCo continuous-control tasks used in our evaluation. (a--c)~Adroit dexterous manipulation tasks using the 24-DoF ShadowHand. (d)~AntMaze navigation task (top-down view).}
        \label{fig:mujoco_tasks}
    \end{figure}

}{}

\subsection{Atari}

For the Atari environments, we build on the standard DeepMind \textsc{Acme} preprocessing wrappers. The environment is initialized using the \texttt{NoFrameskip} variant (or \texttt{v0} with sticky actions) to obtain raw 60\,Hz frames with the full 18-action space exposed.

\paragraph{Temporal pooling and grayscaling}
After each action repeat of 4 frames, temporal max-pooling is applied element-wise over the last 2 of the 4 frames to prevent sprite flickering artifacts. The resulting RGB image is converted to grayscale via the standard luminosity formula $Y = 0.299R + 0.587G + 0.114B$.

\paragraph{Downsampling}
The grayscale image is resized to the target resolution of $84 \times 84$ pixels. Two variants are used:
\begin{itemize}[nosep, leftmargin=*]
    \item \emph{Standard (Montezuma, Pitfall):} The grayscale frame ($210 \times 160$) is directly resized to $84 \times 84$ using bilinear interpolation.
    \item \emph{With spatial max-pooling (Venture):} A $2 \times 2$ spatial max-pooling step is applied first to preserve small, bright sprites (e.g., the player dot) that would otherwise be smoothed out by bilinear interpolation. This reduces the resolution to $105 \times 80$, which is then resized to $84 \times 84$ via bilinear interpolation.
\end{itemize}

\begin{figure}[ht]
    \centering
    \newcommand{\obsimg}[1]{\raisebox{-0.5\height}{\includegraphics[width=0.14\linewidth, trim=0pt 0pt 0pt 18pt, clip]{figures/#1}}}
    \begin{tabular}{@{}m{4em} cccc c@{}}
                          & \textbf{Frame $-3$}     & \textbf{Frame $-2$}     & \textbf{Frame $-1$}     & \textbf{Frame $0$}      & \textbf{Frame top}     \\[2pt]
        \texttt{door}     & \obsimg{door-3.png}     & \obsimg{door-2.png}     & \obsimg{door-1.png}     & \obsimg{door-0.png}     &                        \\[2pt]
        \texttt{hammer}   & \obsimg{hammer-3.png}   & \obsimg{hammer-2.png}   & \obsimg{hammer-1.png}   & \obsimg{hammer-0.png}   &                        \\[2pt]
        \texttt{relocate} & \obsimg{relocate-3.png} & \obsimg{relocate-2.png} & \obsimg{relocate-1.png} & \obsimg{relocate-0.png} &                        \\[2pt]
        \texttt{antmaze}  & \obsimg{antmaze-3.png}  & \obsimg{antmaze-2.png}  & \obsimg{antmaze-1.png}  & \obsimg{antmaze-0.png}  & \obsimg{antmaze-4.png} \\
    \end{tabular}
    \caption{Processed visual observations as seen by the agent for each MuJoCo task. Each column shows a frame in the observation stack, from the oldest (Frame~$-3$) to the most recent (Frame~$0$). For AntMaze, the ``Frame top'' column shows the global top-down view of the maze; during Phase~I (exploration), only this top-down view is used. See \cref{sec:expsetup} for an overview and \cref{app:obs_processing} for full details on the observation processing pipeline.}
    \label{fig:obs_processing}
\end{figure}

\paragraph{Normalization and frame stacking}
Pixel values are cast to floating-point and scaled to $[0, 1]$. The 4 most recent processed frames are stacked along the channel axis, producing a final observation tensor of shape $84 \times 84 \times 4$. For Phase~II (backward learning) on \montezuma, a fifth frame is appended corresponding to the most recent reward observation---i.e., the frame at which the agent last received a reward (see \cref{app:backward} for details).

\subsection{Adroit}

\paragraph{Pixel rendering}
Rather than using the standard \texttt{gym} rendering interface, we attach a \texttt{mujoco.MovableCamera} to the physics state and render directly at $120 \times 120$ pixels in RGB. For each task, the camera pose is locked to a fixed, task-specific viewpoint:
\begin{itemize}[nosep, leftmargin=*]
    \item \texttt{hammer-v0}: Lookat $= [0, {-0.15}, 0.15]$, Distance $= 0.7$, Azimuth $= {-45}^{\circ}$, Elevation $= {-45}^{\circ}$.
    \item \texttt{door-v0}: Lookat $= [0, {-0.1}, 0.25]$, Distance $= 0.8$, Azimuth $= 60^{\circ}$, Elevation $= {-35}^{\circ}$.
    \item \texttt{relocate-v0}: Lookat $= [0, {-0.1}, 0.15]$, Distance $= 0.9$, Azimuth $= {-180}^{\circ}$, Elevation $= {-45}^{\circ}$.
\end{itemize}

\paragraph{Grayscaling, normalization, and frame stacking}
The rendered RGB image is converted to grayscale via the luminosity formula. Four consecutive frames are stacked along the channel axis, producing a tensor of shape $120 \times 120 \times 4$. Pixel values are normalized to $[0, 1]$.

\subsection{AntMaze}

\paragraph{Scene modification}
To make the task visually solvable, the environment scene is dynamically modified: a semi-transparent red sphere is injected at the goal coordinates via a scene callback, and the floor color is set to dark gray ($[0.2, 0.2, 0.2]$) for improved contrast when textures are disabled.

\paragraph{Dual camera rendering}
Two views are rendered at $120 \times 120$ pixels using a \texttt{mujoco.MovableCamera}:
\begin{itemize}[nosep, leftmargin=*]
    \item \emph{Global top-down view:} Locked high above the maze center (Lookat $= [18, 12, 0]$, Distance $= 55$, Azimuth $= 90^{\circ}$, Elevation $= {-90}^{\circ}$). Textures are disabled to provide a clean structural map of the maze layout and goal marker.
    \item \emph{Third-person egocentric view:} Dynamically tracks the agent's position (Distance $= 8$, Azimuth $= 135^{\circ}$, Elevation $= {-45}^{\circ}$) with textures enabled for rich visual feedback of the ant's limbs and surroundings.
\end{itemize}

Both views are converted to grayscale and resized to $120 \times 120$ using Lanczos resampling.

\paragraph{Frame stacking variants}
During Phase~I (exploration), only the global top-down view is used, yielding an observation of shape $120 \times 120 \times 1$. During Phase~II (policy distillation), the global frame is stacked with 4 historical egocentric frames, producing an observation of shape $120 \times 120 \times 5$ (channel 0: global map; channels 1--4: temporal egocentric history). Pixel values are normalized to $[0, 1]$.

\section{MuJoCo Reward and Dead-State Extraction}
\label{app:mujoco_rewards}

This appendix details the sparse reward and dead-state detection logic used during Phase~I exploration for the MuJoCo environments.

\subsection{AntMaze: Flip Detection}
If the ant flips over, it cannot recover, rendering the remainder of the episode useless. We detect flips using two quantities extracted from the MuJoCo physics state at every step:
\begin{enumerate}[nosep, leftmargin=*]
    \item \emph{Z-coordinate} ($z$): the absolute height of the ant torso, retrieved via \texttt{physics.data.qpos[2]}.
    \item \emph{Upward vector} ($z_{\mathrm{up}}$): the $(2,2)$ element of the torso's $3 \times 3$ orientation rotation matrix \[ \texttt{physics.data.xmat[torso\_id]},\]
          representing the Z-component of the torso's local ``up'' vector in the world frame.
\end{enumerate}
The ant is flagged as flipped if either $z < 0.2$ (torso near the ground) or $z_{\mathrm{up}} < 0$ (torso upside-down). Upon detection, the particle is assigned a \dead state. If the agent triggers \texttt{goal\_achieved} on the same step as a flip, the particle is still marked as \dead.

\subsection{Adroit: Sparse Contact Rewards}

For all Adroit tasks, the standard continuous rewards are replaced by a sparse reward wrapper that parses the MuJoCo contact information at every step. Contact is evaluated by extracting geometry (geom) IDs from the model and iterating through active collisions in \texttt{physics.data.contact}, using explicit geom whitelists to prevent false positives (e.g., the hand contacting the table rather than the target object). If the underlying environment flags \texttt{goal\_achieved = True}, the agent receives a $+1$ terminal reward. The task-specific logic is:

\begin{itemize}[leftmargin=*, nosep]
    \item \texttt{hammer-v0}: Contact is checked between the 19 geoms comprising the robot hand (wrist, palm, and finger joints) and the hammer object. A reward of $+1$ is assigned when the hand first contacts the hammer. If contact is subsequently lost, the particle is marked as \dead.
    
    \item \texttt{door-v0}: Contact is checked between the hand geoms and the door latch/handle geoms. A reward of $+1$ is assigned upon first contact. Loss of contact triggers a \dead state.

    \item \texttt{relocate-v0}: Contact is checked between the ball and all geoms \emph{except} a whitelist of forbidden surfaces (floor and table geoms). A reward of $+1$ is assigned when any hand surface first contacts the ball. If contact with the ball is subsequently lost, the particle is marked as \dead.
\end{itemize}

    \section{Ablation Study}
    \label{sec:ablations}

    \begin{figure}[!htbp]
        \centering
          \resizebox{\columnwidth}{!}{\begin{tabular}{ccc}
        \includegraphics[width=0.32\linewidth]{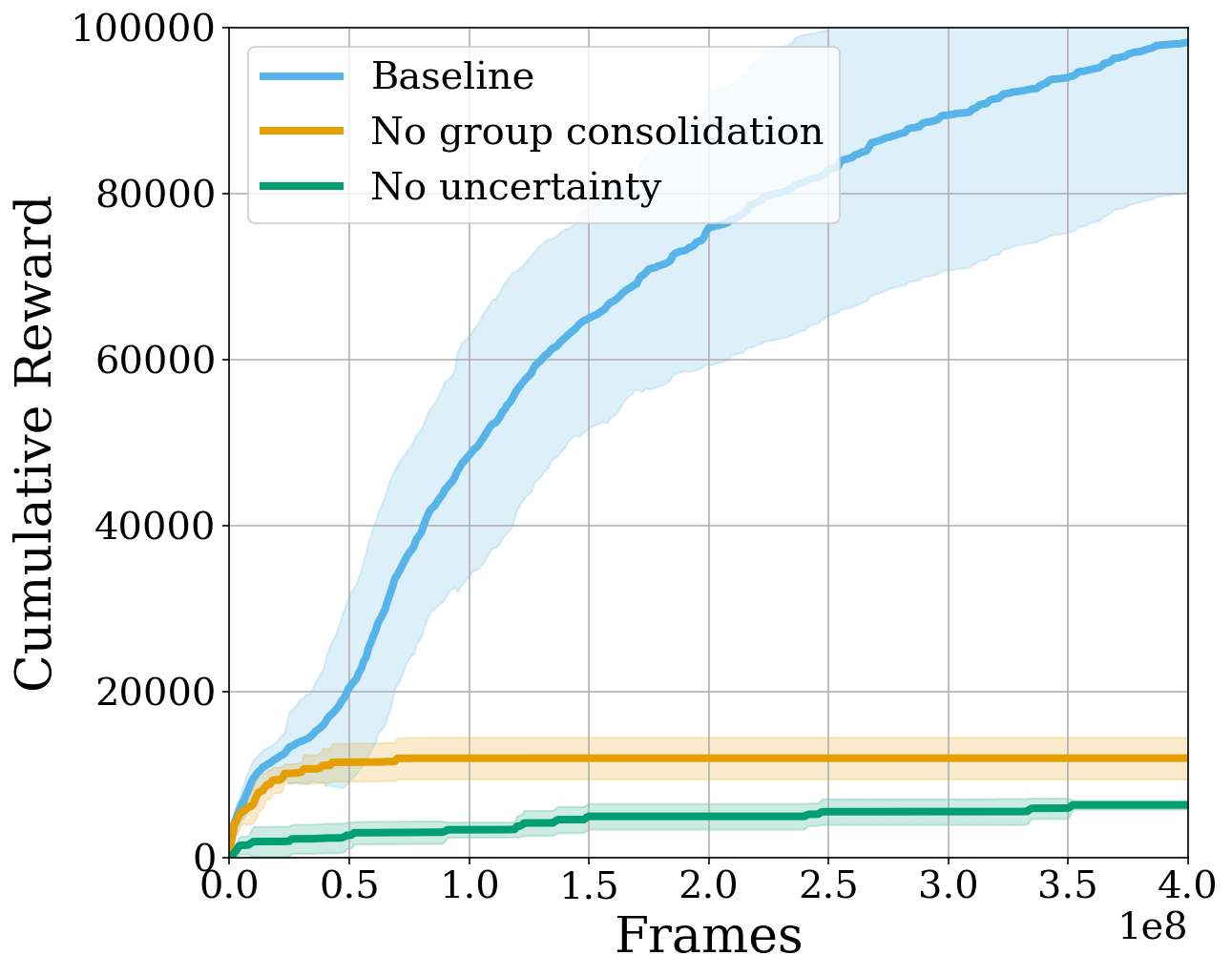} &
        \includegraphics[width=0.32\linewidth]{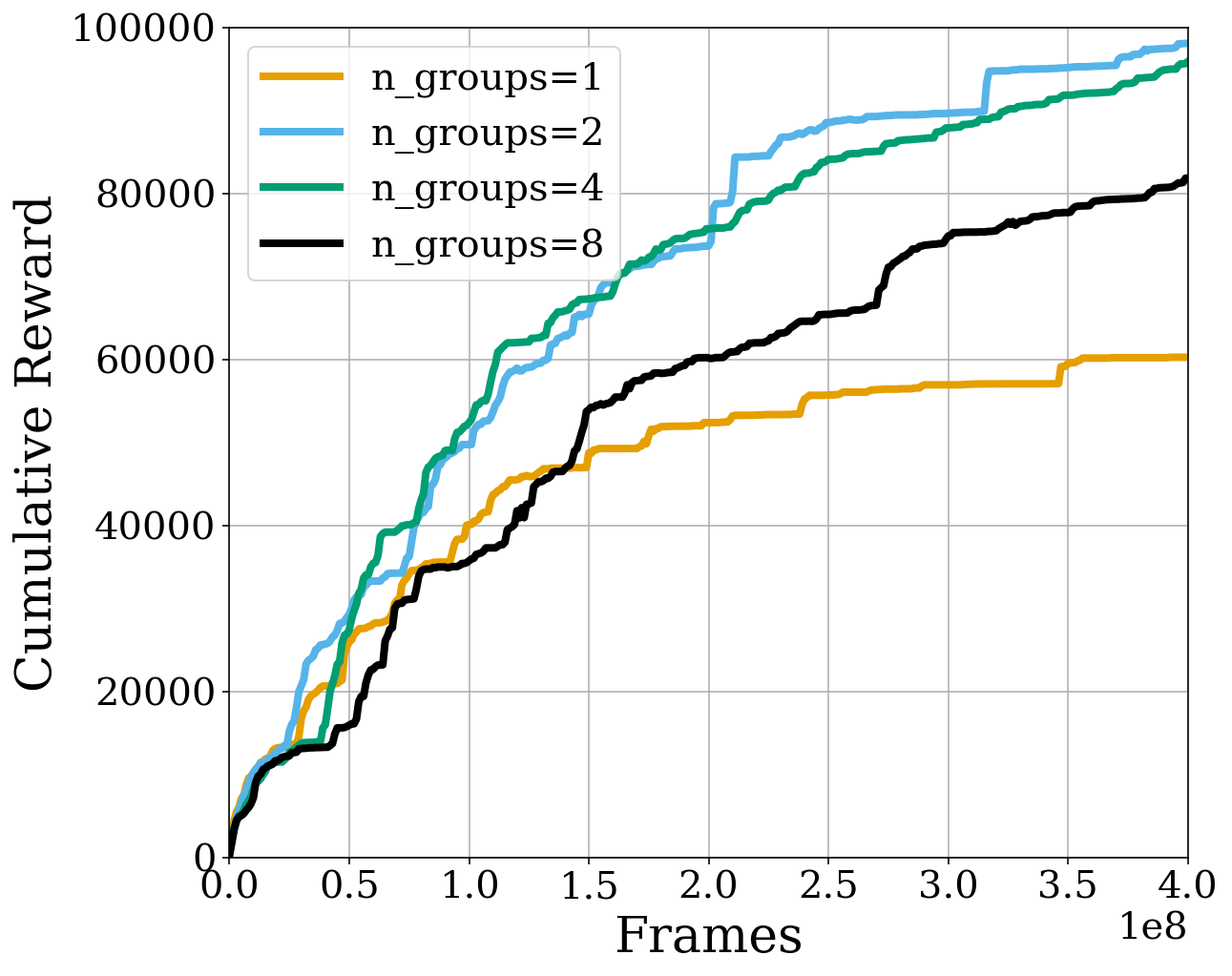} &
        \includegraphics[width=0.32\linewidth]{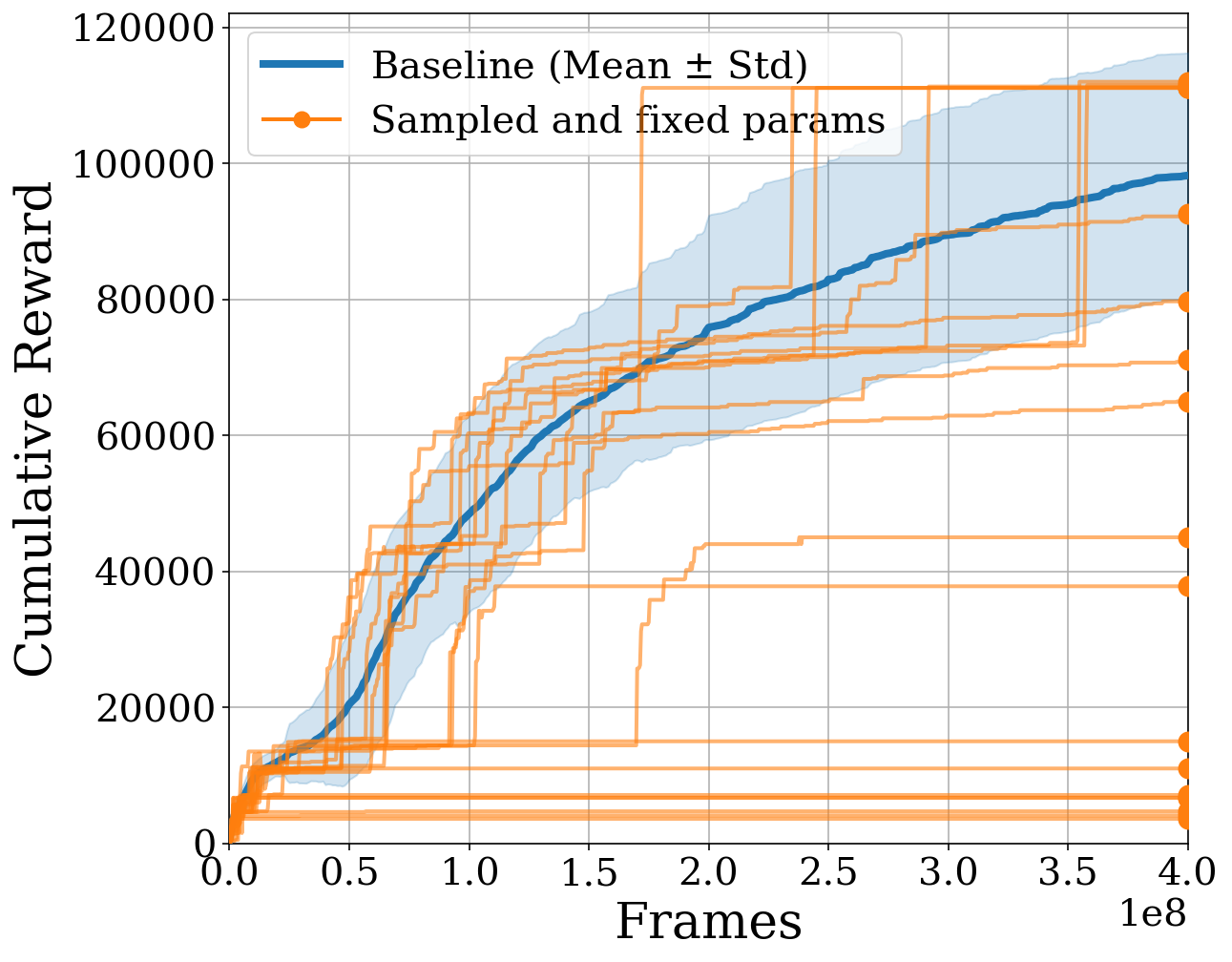} \\
        {\small (a) Component ablation} & {\small (b) Effect of group size} & {\small (c) Fixed vs.\ randomized hyperparameters}
        \end{tabular}
          }
        \caption{Ablation studies on \montezuma. Mean cumulative reward ($\pm$ std) across seeds. (a)~Disabling uncertainty-based winner selection causes exploration to fail; disabling group consolidation degrades performance. (b)~$2$ and $4$ groups perform similarly; $1$ group is worst; $8$ groups improve over $1$ but underperform $2$ and $4$. (c)~Per-iteration hyperparameter randomization yields robust performance without tuning.}
        \label{fig:ablations}
    \end{figure}

    To validate the design choices underlying \gowu, we conduct a series of ablation experiments on \montezuma. Unless stated otherwise, we report the mean cumulative reward (with standard deviation) along the exploration path as a function of game frames.

    All ablations are conducted on \montezuma with $8$ seeds per variant (the baseline uses $100$ seeds; see \cref{sec:results}). \cref{fig:ablations} presents the results across three experiments.

    \paragraph{Component ablation}
    We disable individual components of \gowu while keeping everything else fixed (\cref{fig:ablations}a). Removing the uncertainty estimator (selecting the winner uniformly at random among surviving particles instead) causes exploration to fail; the variant does not complete the first level. This confirms that epistemic uncertainty is critical for directing the search toward under-explored regions. Disabling the group consolidation step (\cref{line:single}) also degrades performance, though the variant still does slightly better than the no-uncertainty ablation. This indicates that periodically collapsing the population to the group's most uncertain state accelerates progress. An additional benefit of group consolidation is that it results in more frequent pruning, keeping the lineage tree smaller and reducing memory usage.

    \paragraph{Effect of group size}
    We vary the number of parallel groups $M \in \{1, 2, 4, 8\}$ while keeping the total number of particles fixed at $128$ (i.e., $N = 128/M$ per group) (\cref{fig:ablations}b). Using $M = 2$ and $M = 4$ groups yields similar performance; $M = 1$ (a single group) is the worst, and $M = 8$ improves over $M = 1$ but underperforms $M \in \{2, 4\}$, because the number of particles per group drops to $16$, making it harder to leverage the particle management logic (winner selection, pruning, and rollback) to navigate around obstacles.
 
    \paragraph{Robustness to hyperparameter randomization}
    In our standard configuration, the population management hyperparameters ($K$, $T$, and rollback depth) are sampled uniformly from their respective ranges at each iteration. To test robustness, we run an alternative protocol with $20$ seeds in which a single draw from each range is fixed for the entire run; different seeds thus correspond to different fixed hyperparameter settings (\cref{fig:ablations}c). Per-iteration randomization yields consistently strong performance, whereas the fixed-draw protocol produces high variance across seeds. This demonstrates that randomization provides inherent robustness: sampling from a reasonable range at each step is sufficient, without needing to identify a single optimal setting.

\section{Additional Related Works}
\label{app:additional_related}

\paragraph{Evolutionary and population-based methods}
\gowu's population-based nature connects it to evolutionary methods, which drive exploration by evolving diverse populations of policies. Foundational approaches such as Novelty Search, MAP-Elites, and the quality-diversity literature reward behavioral novelty or coverage \citep{lehman2011abandoning,mouret2015illuminating,pugh2016quality}. More recent hybrids combine such diversity mechanisms with deep RL and policy-gradient updates \citep{conti2018improving,parkerholder2020effective,wang2025evolutionary}. Although both families maintain populations, their mechanisms for driving exploration differ fundamentally: unlike \gowu, which searches directly in state space, these methods operate in the parameter space of a policy, generating diversity by perturbing model weights.

\paragraph{Additional differences between \gowu and MCTS}
\gowu and MCTS also differ in how search is guided and in the kinds of action spaces they most naturally accommodate. Standard MCTS is usually guided by value estimates or visit counts, whereas \gowu uses epistemic uncertainty as a primary signal for redistributing exploration effort. Recent work on Epistemic MCTS augments MCTS with epistemic uncertainty for deeper exploration \citep{oren2025epistemic}, but remains a non-particle-based search method. In addition, standard MCTS most naturally fits discrete action spaces, while continuous-action variants typically require extra machinery such as progressive widening \citep{couetoux2011continuous}. \gowu does not enumerate actions at tree nodes in this way: particles can execute rollouts using arbitrary policies, which allows the framework to accommodate continuous actions more directly.

\paragraph{Latent Go-Explore}
Latent Go-Explore \citep{gallouedec2023cellfree} addresses the reliance of Go-Explore on hand-designed observation discretization by replacing it with a learned latent representation. However, learning such a representation itself requires sufficiently rich exploration data, creating a potential chicken-and-egg issue. Moreover, the method is evaluated only in terms of exploration performance, where it slightly improves over Go-Explore on \montezuma and \pitfall. It remains unclear whether this cell-free variant also supports the subsequent policy-learning stage needed to obtain a deployable policy.
 
\paragraph{MuJoCo baselines}
For the MuJoCo tasks evaluated in this work, direct baselines are unavailable.

For Adroit, recent visual RL methods such as \textsc{DrM} \citep{xu2023drm} and \textsc{MENTOR} \citep{huang2024mentor} solve \texttt{door} and \texttt{hammer} from pixels without demonstrations, but rely on dense reward shaping. With privileged state observations, \citet{wang2024learning} solve \texttt{door} and \texttt{hammer} using sparse rewards and model-based intrinsic motivation, but do not use pixels. To the best of our knowledge, no existing method solves these tasks from \emph{pixel observations} in the sparse-reward setting without expert demonstrations or offline datasets. The \texttt{relocate} task, in particular, remains unsolved from pixels even with dense rewards; with state observations, it has been solved using dense reward shaping \citep{rajeswaran2017learning}. We note that our Phase~I exploration does augment the sparse environment reward with a single intermediate reward signal and a dead-state condition, both derived from privileged state information (see \cref{app:mujoco_rewards} for details); these are far simpler than the dense reward shaping required by standard RL methods, but they do go beyond a purely sparse reward signal.

For AntMaze, \textsc{Director} \citep{hafner2022director} solves an egocentric ant maze from first-person pixel inputs with sparse rewards, but their custom maze is smaller than ours (\texttt{antmaze-large-diverse-v0}) and uses uniquely colored walls as navigation landmarks. Several goal-conditioned RL methods \citep{bortkiewicz2024accelerating,kim2023landmark} achieve moderate success rates on AntMaze benchmarks using the JaxGCRL framework (e.g., $\sim\!65\%$ on AntMaze-large), but operate from state observations with fixed start and goal locations. In contrast, our setup learns entirely from images with randomized goal locations. We note, however, that our training uses the default environment randomization, where the target location is slightly perturbed around a fixed point across seeds. This differs from the D4RL-style ``diverse'' setting \citep{fu2020d4rl}, where trajectories span substantially different start and goal configurations, requiring greater generalization.

\paragraph{Resets and simulation in RL}
Resets are sometimes viewed as infeasible for physical robots \citep{eysenbach2017leavetracelearningreset,gupta2021resetfreereinforcementlearningmultitask}, but this view overlooks the central role of simulation in modern AI training. Even for physical applications, policies are increasingly pre-trained in simulation---from robotic manipulation to autonomous driving \citep{qassem2010modeling,bojarski2016end,tobin2017domainrandomizationtransferringdeep,akkaya2019solving,bansal2018chauffeurnetlearningdriveimitating}---making resets a readily available primitive. The empirical success of planning algorithms such as AlphaZero \citep{silver2016mastering,silver2018general} provides further evidence for the power of reset-based search.

\paragraph{Temporal contrastive features for exploration}
Temporal contrastive features \citep{myers2024learning,mohamed2026temporalrepresentationsexplorationlearning,liu2024single} learn representations that capture the temporal structure of the environment, offering a potentially stronger basis for planning-aware and noise-robust exploration than prediction-error-based methods such as \rnd.

\section{\montezuma Middle Room Bug}
\label{app:middleroombug}

\begin{figure}[h] 
\centering
\setlength{\tabcolsep}{4pt}
\resizebox{\columnwidth}{!}{\begin{tabular}{@{}ccccc@{}}
    \includegraphics[height=3.0cm, trim=0pt 0pt 0pt 18pt, clip]{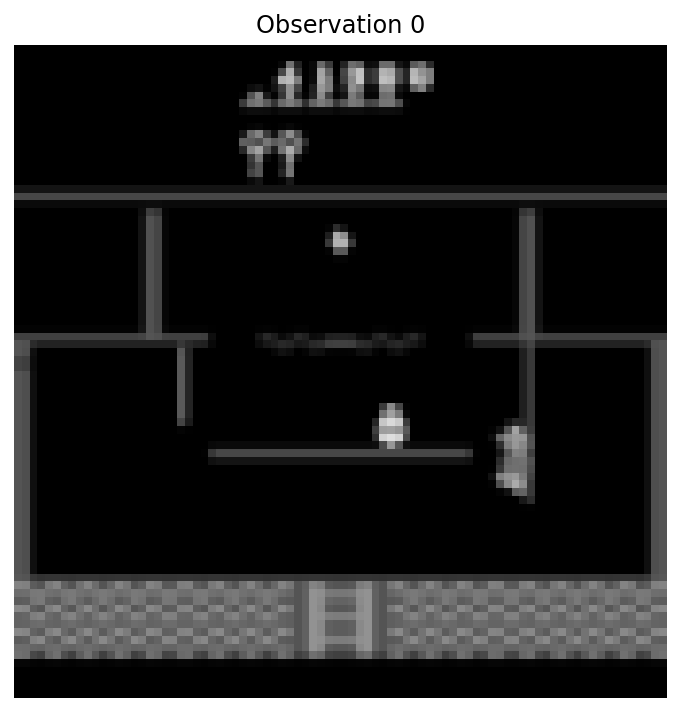} &
    \includegraphics[height=3.0cm, trim=0pt 0pt 0pt 18pt, clip]{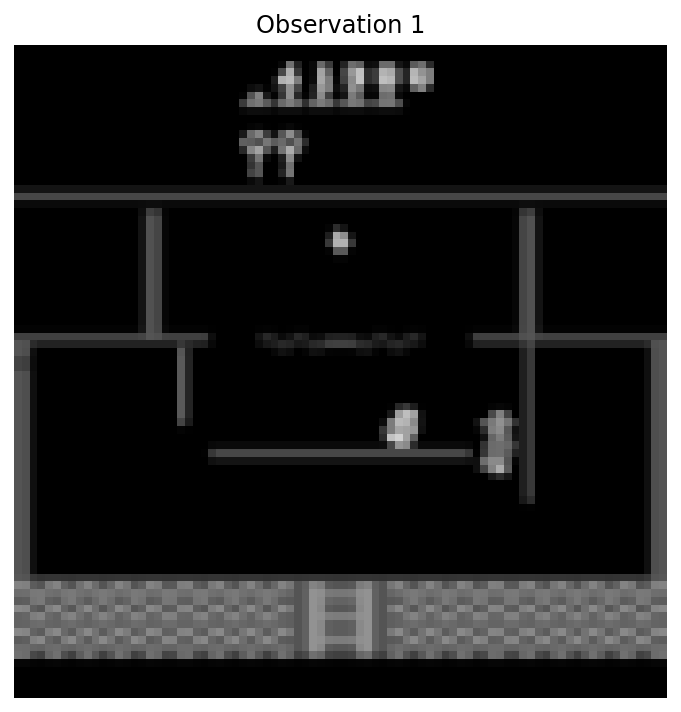} &
    \includegraphics[height=3.0cm, trim=0pt 0pt 0pt 18pt, clip]{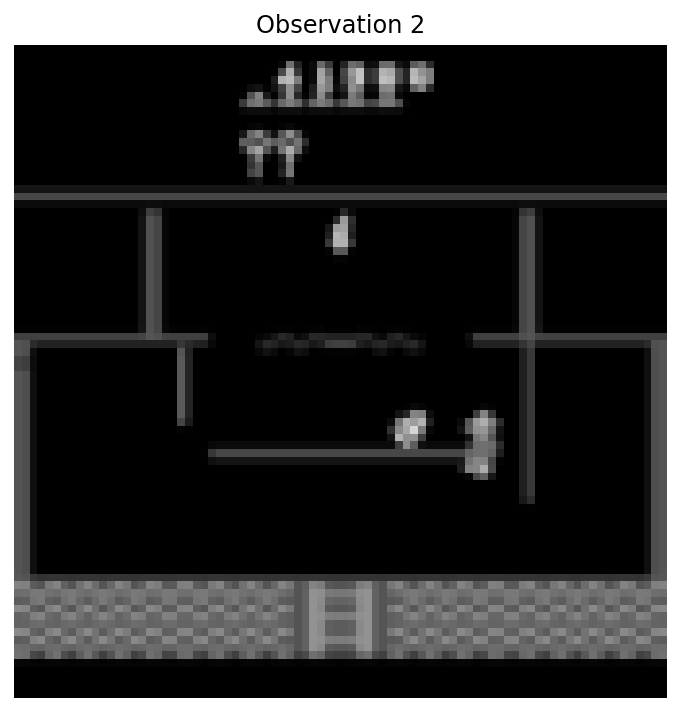} &
    \includegraphics[height=3.0cm, trim=0pt 0pt 0pt 18pt, clip]{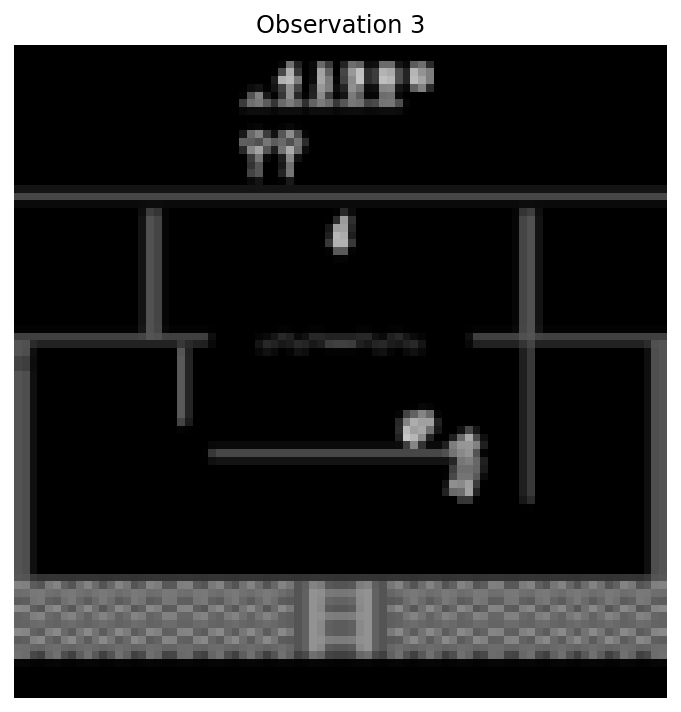} &
    \includegraphics[height=3.0cm, trim=0pt 0pt 0pt 18pt, clip]{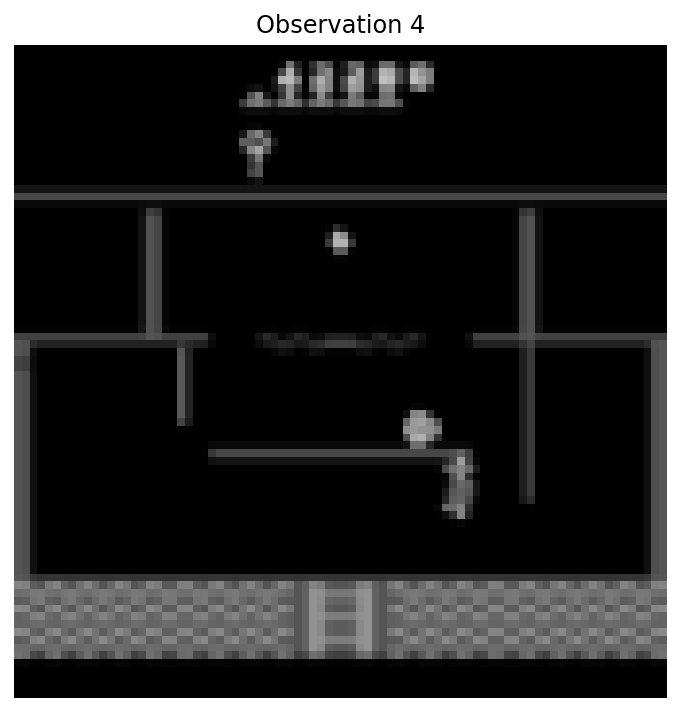} \\[4pt]
    {\small Frame $-4$} & {\small Frame $-3$} & {\small Frame $-2$} & {\small Frame $-1$} & {\small Frame $0$}
\end{tabular}
}
\caption{Frame-by-frame illustration of the middle room bug in \montezuma. The agent starts on the rope (Frame~$-4$), jumps left, and hits the skull platform (Frame~$0$), losing a key and collecting an unintended reward.} 
\label{fig:middleroombug}
\end{figure}

During exploration, we discovered a previously undocumented bug in \montezuma that occurs in the \emph{middle room} of each level (the room containing a torch and a rolling skull). \cref{fig:middleroombug} illustrates the bug sequence frame by frame: the agent, positioned on the rope on the right-hand side of the room (Frame~$-4$), jumps to the left and progressively approaches the platform where the rolling skull patrols (Frames~$-3$ through~$-1$), eventually making contact with the platform (Frame~$0$). If the agent is carrying a key at the time of contact, the key is consumed and a reward is collected. Since the key in the first room respawns periodically after a sufficient delay \citep{salimans2018learning}, the agent can return, re-acquire it, and repeat the process indefinitely, creating an infinite reward loop that traps the agent in the same level and prevents further exploration.

To prevent the algorithm from exploiting this glitch, we employ a two-step visual detection heuristic that activates whenever a positive reward is registered. 

\paragraph{Step 1: Location verification}
We first determine whether the agent is currently in the middle room. A fixed $20 \times 20$ spatial crop of the current frame is extracted from a region that uniquely identifies this room. The crop is compared against a pre-recorded set of reference crops from the middle room across different game levels using the L2 norm (Euclidean distance). If the pixel-wise distance is below a small threshold, the agent is confirmed to be in the middle room.

\paragraph{Step 2: Action verification}
If the location check passes, we verify whether the agent has just performed the bugged action (jumping off the rope). We inspect the agent's last four frames (as shown in \cref{fig:middleroombug}) and isolate a bounding box covering the area where the rope sequence takes place. The maximum pixel-wise difference between consecutive frames in this region is computed; if it exceeds a threshold, it indicates a sudden burst of movement consistent with the visual signature of the agent jumping off the rope.

\paragraph{Intervention}
If a positive reward is received while \emph{both} visual conditions are simultaneously satisfied, the reward is flagged as an exploit and the particle is immediately marked as \dead. This ensures the buggy particle will never be selected as a winner; instead, at the next \gowu redistribution step, it will be replaced by a clone of the current winner---a non-buggy particle---and exploration resumes from a valid state.

\section{\montezuma Key Respawn Mechanism}
\label{app:keyrespawn}

In \montezuma, the key in the first room respawns after a sufficient amount of time has elapsed, a feature previously documented by \citet{salimans2018learning}. Some of our highest-scoring distilled policies exploit this mechanism: rather than traveling to distant rooms to collect keys, they learn to move back and forth between the first room and an adjacent room, waiting for the key to reappear. By collecting keys without traversing the full level, the agent can sometimes complete levels faster, leading to the especially high scores observed for the best Phase~II runs. We note that this is a feature of the game, not a bug; in fact, waiting for the key to respawn is sometimes necessary, as certain levels require the first room's key multiple times to advance.

\section{Why GWTW is Exponentially Faster: A Concrete Example}
\label{app:gwtw_examples}

Consider the problem of finding the deepest node in an unknown tree. DFS and BFS may, in the worst case, need to visit every node before reaching it. \gwtw takes a different approach: it advances a population of particles in parallel, kills those that reach leaves, and clones survivors to maintain the population size. This pruning-and-cloning mechanism redirects all computational effort toward branches that remain alive. We illustrate this with a concrete tree where DFS and BFS both require $\Theta(2^D)$ node evaluations to find the unique deepest node, while \gwtw succeeds with only $\mathcal{O}(D \log(D/\delta))$.

\paragraph{Tree construction}
Let $D$ be a depth parameter. We construct a tree $T$ with a unique deepest node at depth $D+1$:
\begin{itemize}[nosep, leftmargin=*]
    \item \textbf{Spine (golden path):} Nodes $v_0$ (root), $v_1, \ldots, v_{D+1}$. Each spine node $v_i$ ($0 \le i \le D-1$) has exactly two children: $v_{i+1}$ and a trap root $t_i$. Node $v_D$ has a single child $v_{D+1}$, which is a leaf and the unique deepest node.
    \item \textbf{Trap subtrees:} Each $t_i$ roots a subtree where every internal node has $6$ children: $2$ internal nodes and $4$ leaves (immediate dead ends). Trap subtrees extend down to depth~$D-1$, at which point all $6$ children are leaves (at depth $D$).
\end{itemize}
Because each trap node spawns $2$ internal children, the traps grow like binary trees. The total number of nodes in $T$ is $\Theta(2^D)$.

\paragraph{Inefficiency of DFS and BFS}
Both algorithms must process the full tree volume:
\begin{itemize}[nosep, leftmargin=*]
    \item \textbf{BFS} explores layer by layer. To reach depth $D+1$, it must visit all $\Theta(2^D)$ nodes at preceding levels. \textbf{Cost: $\Theta(2^D)$.}
    \item \textbf{DFS} (randomized) has a $1/2$ probability of entering a trap at each spine node. Since the earliest traps contain $\Theta(2^D)$ nodes that must be exhaustively explored before backtracking, even a single wrong turn is catastrophic. \textbf{Cost: $\Theta(2^D)$.}
\end{itemize}
Note that independent random walks also fail: reaching $v_{D+1}$ requires taking the correct branch $D$ times in a row, which happens with probability $2^{-D}$.

\paragraph{Analysis of \gwtw}
\gwtw maintains $B$ particles, all starting at~$v_0$. At each step: (1)~every particle moves to a uniformly random child; (2)~particles landing on leaves die; (3)~survivors are resampled with replacement to restore the population to $B$.

Let $x_i$ denote the expected fraction of particles at spine node $v_i$. When particles at $v_i$ step forward, both children ($v_{i+1}$ and $t_i$) are non-leaves, so all golden-path particles survive: $x_i/2$ land on $v_{i+1}$ and $x_i/2$ enter the trap. Meanwhile, trap particles each choose among $6$ children ($2$ internal, $4$ leaves), surviving with probability $1/3$. The total survivor fraction is:
\begin{equation}
    S_i \;=\; \underbrace{x_i}_{\text{golden survivors}} + \; \underbrace{\frac{1-x_i}{3}}_{\text{trap survivors}} \;=\; \frac{1 + 2x_i}{3}.
\end{equation}
After resampling, the fraction on the golden path becomes:
\begin{equation}
    x_{i+1} \;=\; \frac{x_i/2}{S_i} \;=\; \frac{3x_i}{2(1 + 2x_i)}.
    \label{eq:gwtw_recurrence}
\end{equation}
Starting from $x_0 = 1$, this recurrence produces: $1 \to 1/2 \to 3/8 \to 9/28 \to \cdots$, converging monotonically from above to the fixed point $x^* = 1/4$. Since the map $f(x) = \frac{3x}{2+4x}$ is strictly increasing with $f(1/4) = 1/4$, the sequence remains above $1/4$ for all $i$. This is the key insight: because traps kill particles at rate $2/3$ per step while the golden path kills none, \gwtw automatically maintains at least $25\%$ of its population on the correct path, regardless of the exponential volume of the traps.

\paragraph{Cost of \gwtw}
At each depth, the expected number of golden-path particles is at least $B/4$. By a Chernoff bound, the probability that this count drops to zero at any single depth is at most $\exp(-cB)$ for a constant $c > 0$. Applying a union bound over $D$ levels, the probability that \gwtw ever loses the golden path is at most $D \cdot \exp(-cB) \le \delta$ whenever $B \ge \frac{1}{c}\ln \big(\frac{D}{\delta}\big)$. The total number of node evaluations is therefore:
\begin{equation}
    \mathrm{Cost}_{\textsc{gwtw}} \;=\; \mathcal{O} \left(B \cdot D\right) \;=\; \mathcal{O}\left(D \log \frac{D}{\delta}\right),
\end{equation}
which is exponentially smaller than the $\Theta(2^D)$ cost of DFS and BFS. Note that for any fixed search strategy, adversarial trees can be constructed that force $\Omega(2^D)$ work. The fundamental reason \gwtw succeeds is that parallel exploration with redistribution amplifies the probability of advancing at each depth level: in the example above, at each level \gwtw only requires \emph{at least one} of $B$ particles to follow the spine rather than enter a trap, after which cloning restores the population to $B$ for the next level. A single walker, by contrast, must choose the spine over a trap at all $D$ levels in sequence, succeeding with probability $2^{-D}$. \gwtw converts this multiplicative success probability into an additive failure probability of $D \cdot e^{-cB}$ for a constant $c > 0$.

\neurips{
  \newpage
  \section*{NeurIPS Paper Checklist}

\begin{enumerate}

\item {\bf Claims}
    \item[] Question: Do the main claims made in the abstract and introduction accurately reflect the paper's contributions and scope?
    \item[] Answer: \answerYes{}
    \item[] Justification: The abstract and introduction claim three contributions: (1) a new exploration paradigm that bypasses intrinsic-reward policy optimization, (2) state-of-the-art results on hard-exploration Atari games, and (3) solving MuJoCo Adroit and AntMaze tasks from pixels with sparse rewards. All three are supported by the experimental results in \cref{sec:experiments} and the appendix.

\item {\bf Limitations}
    \item[] Question: Does the paper discuss the limitations of the work performed by the authors?
    \item[] Answer: \answerYes{}
    \item[] Justification: The Discussion and Future Work section (\cref{sec:discussion}) discusses limitations including the reliance on the reset primitive (which limits applicability to physical systems), the use of \rnd as the uncertainty estimator (which can assign high novelty to stochastic observations), and the need for environment-specific dead-state definitions. The appendix (\cref{app:additional_related}) further qualifies the MuJoCo claims, noting differences from prior setups.

\item {\bf Theory assumptions and proofs}
    \item[] Question: For each theoretical result, does the paper provide the full set of assumptions and a complete (and correct) proof?
    \item[] Answer: \answerNA{}
    \item[] Justification: The paper does not claim formal theoretical results. The analysis in \cref{app:gwtw_examples} provides an illustrative example with a cost analysis of GWTW on a specific tree construction, but this is presented as a motivating example rather than a formal theorem.

    \item {\bf Experimental result reproducibility}
    \item[] Question: Does the paper fully disclose all the information needed to reproduce the main experimental results of the paper to the extent that it affects the main claims and/or conclusions of the paper (regardless of whether the code and data are provided or not)?
    \item[] Answer: \answerYes{}
    \item[] Justification: The paper provides complete algorithmic pseudocode (\cref{alg:population_iteration,alg:parallel_evolution}), all hyperparameters (\cref{tab:hyperparams,tab:backward_config}), model architectures (\cref{app:uncertaintyrnd,app:backward}), distributed system design (\cref{app:implementation}), observation processing pipelines (\cref{app:obs_processing}), environment-specific configurations (\cref{app:backward_envconfig}), and checkpointing details (\cref{app:checkpointing}).

\item {\bf Open access to data and code}
    \item[] Question: Does the paper provide open access to the data and code, with sufficient instructions to faithfully reproduce the main experimental results, as described in supplemental material?
    \item[] Answer: \answerNo{}
    \item[] Justification: Code is not released at submission time but will be made publicly available at a later date. All algorithmic details, hyperparameters, and implementation specifics are fully documented in the paper and appendix to support independent reproduction.

\item {\bf Experimental setting/details}
    \item[] Question: Does the paper specify all the training and test details (e.g., data splits, hyperparameters, how they were chosen, type of optimizer) necessary to understand the results?
    \item[] Answer: \answerYes{}
    \item[] Justification: Phase~I hyperparameters are listed in \cref{tab:hyperparams}; Phase~II PPO hyperparameters and backward algorithm configuration are detailed in \cref{app:backward}; the distributed architecture is described in \cref{app:implementation}; observation processing pipelines are specified per environment in \cref{app:obs_processing}; computational resources are reported in \cref{tab:compute}; and environment-specific reward and dead-state logic is documented in \cref{app:mujoco_rewards}.

\item {\bf Experiment statistical significance}
    \item[] Question: Does the paper report error bars suitably and correctly defined or other appropriate information about the statistical significance of the experiments?
    \item[] Answer: \answerYes{}
    \item[] Justification: All main results report mean $\pm$ standard deviation. Phase~I exploration results are averaged over 100 seeds (\cref{tab:main_comparison}); Phase~II policy scores are averaged over 10 runs, each repeated with 5 random seeds (\cref{tab:robustification_stats,tab:mujoco_robustification_stats}). Exploration curves in \cref{fig:atari_exploration} show mean $\pm$ std shading. MuJoCo exploration statistics include mean, std, median, min, and max (\cref{tab:mujoco_exploration}).

\item {\bf Experiments compute resources}
    \item[] Question: For each experiment, does the paper provide sufficient information on the computer resources (type of compute workers, memory, time of execution) needed to reproduce the experiments?
    \item[] Answer: \answerYes{}
    \item[] Justification: \Cref{tab:compute} reports wall-clock runtime and computational resources for Phase~I on \montezuma (1 TPU, 128 CPUs, 1 coordinator CPU). Phase~II training budgets (in environment frames) are reported per task in \cref{tab:mujoco_phaseii_budget} and in the main text for Atari (15--20B frames).
    
\item {\bf Code of ethics}
    \item[] Question: Does the research conducted in the paper conform, in every respect, with the NeurIPS Code of Ethics \url{https://neurips.cc/public/EthicsGuidelines}?
    \item[] Answer: \answerYes{}
    \item[] Justification: The research involves algorithmic development and simulation-based experiments using publicly available benchmark environments (Atari, MuJoCo). No human subjects, private data, or sensitive applications are involved.

\item {\bf Broader impacts}
    \item[] Question: Does the paper discuss both potential positive societal impacts and negative societal impacts of the work performed?
    \item[] Answer: \answerNA{}
    \item[] Justification: This work is foundational research on exploration algorithms for reinforcement learning, evaluated in simulation environments (Atari games and MuJoCo). It does not have direct societal applications and we do not foresee specific negative societal impacts.
    
\item {\bf Safeguards}
    \item[] Question: Does the paper describe safeguards that have been put in place for responsible release of data or models that have a high risk for misuse (e.g., pre-trained language models, image generators, or scraped datasets)?
    \item[] Answer: \answerNA{}
    \item[] Justification: The paper does not release pre-trained models, datasets, or other assets that pose a risk for misuse. The method operates in simulation environments and produces task-specific policies for game playing and robotic control.

\item {\bf Licenses for existing assets}
    \item[] Question: Are the creators or original owners of assets (e.g., code, data, models), used in the paper, properly credited and are the license and terms of use explicitly mentioned and properly respected?
    \item[] Answer: \answerYes{}
    \item[] Justification: All benchmark environments used (Arcade Learning Environment, MuJoCo, D4RL) are properly cited. Baseline results are attributed to their original papers with explicit citations.

\item {\bf New assets}
    \item[] Question: Are new assets introduced in the paper well documented and is the documentation provided alongside the assets?
    \item[] Answer: \answerNA{}
    \item[] Justification: The paper does not release new datasets, models, or code assets at this time.

\item {\bf Crowdsourcing and research with human subjects}
    \item[] Question: For crowdsourcing experiments and research with human subjects, does the paper include the full text of instructions given to participants and screenshots, if applicable, as well as details about compensation (if any)? 
    \item[] Answer: \answerNA{}
    \item[] Justification: The paper does not involve crowdsourcing or research with human subjects.

\item {\bf Institutional review board (IRB) approvals or equivalent for research with human subjects}
    \item[] Question: Does the paper describe potential risks incurred by study participants, whether such risks were disclosed to the subjects, and whether Institutional Review Board (IRB) approvals (or an equivalent approval/review based on the requirements of your country or institution) were obtained?
    \item[] Answer: \answerNA{}
    \item[] Justification: The paper does not involve research with human subjects.

\item {\bf Declaration of LLM usage}
    \item[] Question: Does the paper describe the usage of LLMs if it is an important, original, or non-standard component of the core methods in this research? Note that if the LLM is used only for writing, editing, or formatting purposes and does \emph{not} impact the core methodology, scientific rigor, or originality of the research, declaration is not required.
\item[] Answer: \answerNA{}
    \item[] Justification: LLMs are not used as a component of the core methodology in this research.

\end{enumerate} }

\end{document}